%% file: low_rank_medical_imaging.tex
\documentclass[accepted]{uai2024} 

\usepackage[american]{babel}

\hypersetup{
colorlinks=true,citecolor=blue
}

\usepackage{amssymb}
\usepackage{amsmath,mathrsfs,dsfont}
\usepackage{nicefrac}
\usepackage{multirow}
\usepackage[algo2e]{algorithm2e}
\usepackage{algorithmic}
\usepackage{algorithm}

\usepackage{graphicx}
\usepackage{subfigure}
\usepackage{booktabs} 
\usepackage{color}
\usepackage{enumitem}
\usepackage{array}
\usepackage{graphicx,tikz}
\usepackage[mathscr]{euscript}
\usepackage{amsthm}
\usepackage{multirow}

\usepackage{bm}
\usepackage{bbm}
\usepackage{color}
\usepackage{epstopdf}
\usepackage{cleveref}
\usepackage{thmtools}
\usepackage{thm-restate}

\usepackage{bbding}
\usepackage{mathtools}
\usepackage{arydshln}
\allowdisplaybreaks
\usepackage{natbib} 
\usepackage{mathtools} 
\usepackage{tikz} 

\input{symbol_tit.tex}

\newcommand{\KL}{\textup{KL}}

\newcommand{\bWlin}{\bW^{\textup{(lin)}}}

\newcommand{\Net}{\textup{NN}}
\title{Learning Low-Rank Feature for Thorax Disease Classification}

\author[1]{Rajeev Goel\thanks{Indicates equal contribution.}}
\author[1]{Utkarsh Nath$^*$}
\author[1]{Yancheng Wang$^*$}
\author[2]{Alvin C. Silva}
\author[1]{Teresa Wu}
\author[1]{Yingzhen Yang}
\affil[1]{%
    School of Computing and Augmented Intelligence \\
    Arizona State Univerisity\\
    Tempe, AZ 85281, USA \\
    \texttt{\{rgoel15,unath,ywan1053,Teresa.Wu,yingzhen.yang\}@asu.edu}
}
\affil[2]{%
    Department of Radiology\\
    Mayo Clinic Arizona
    Scottsdale, AZ 85259, USA
    \texttt{silva.alvin@mayo.edu}
}

\begin{document}

\maketitle

\begin{abstract}
Deep neural networks, including Convolutional Neural Networks (CNNs) and Visual Transformers (ViT), have achieved stunning success in medical image domain. We study thorax disease classification in this paper. Effective extraction of features for the disease areas is crucial for disease classification on radiographic images. While various neural architectures and training techniques, such as self-supervised learning with contrastive/restorative learning, have been employed for disease classification on radiographic images, there are no principled methods which can effectively reduce the adverse effect of noise and background, or non-disease areas, on the radiographic images for disease classification. To address this challenge, we propose a novel Low-Rank Feature Learning (LRFL) method in this paper, which is universally applicable to the training of all neural networks. The LRFL method is both empirically motivated by the low frequency property observed on all the medical datasets in this paper, and theoretically motivated by our sharp generalization bound for neural networks with low-rank features. In the empirical study, using a neural network such as a ViT or a CNN pre-trained on unlabeled chest X-rays by Masked Autoencoders (MAE), our novel LRFL method is applied on the pre-trained neural network and demonstrate better classification results in terms of both multiclass area under the receiver operating curve (mAUC) and classification accuracy.
\end{abstract}
\section{Introduction}

A chest radiograph~\citep{van2001computer}, commonly known as a chest X-ray, is the predominant diagnostic imaging method in diagnosing abnormal conditions in the airways, blood vessels, bones, heart, lungs, and other structures within the chest cavity. Following the huge success of deep learning in computer vision~\citep{he2016deep, Lin2017a, lin2017feature}, there is a growing interest in developing deep neural networks (DNNs) to detect abnormalities in anatomy in chest X-rays~\citep{guendel2018learning, xiao2023delving, ccalli2021deep, guan2018multi}. The integration of DNNs into radiography practices can assist radiologists in providing more accurate and timely diagnoses.
Accurate clinical decision-making with DNNs heavily relies on learning informative medical feature representation. Early works usually adopt convolutional neural networks (CNNs) such as U-Net~\citep{ronneberger2015u} and its variants~\citep{falk2019u, zhou2018unet++, cui2019pulmonary} for representation learning on radiography images. However, CNNs are biased in detecting local patterns in images, such as edges, shapes, and textures~\citep{chefer2021transformer, bai2021transformers, dosovitskiy2020image, xiao2023delving}.
Different from photograph image processing, long-range feature dependencies are crucial indicators of abnormalities in radiography images~\citep{xie2021cotr, hatamizadeh2022unetr, shamshad2022transformers, xiao2023delving}.
To address that issue, Visual Transformers (ViTs) are adopted to learn more informative medical representations from radiography images~\citep{ma2022benchmarking, chen2021transunet, xiao2023delving}, utilizing their capabilities in capturing long-range feature dependencies via self-attention. Albeit the success of CNNs and ViTs in analyzing radiography images, their accuracy heavily relies on the quality and quantity of data and annotations~\citep{feng2020parts2whole}. However, the collection of large amounts of training data and high-quality annotations in the medical imaging domain is extremely hard~\citep{zhou2021towards}. To tackle this problem, self-supervised learning (SSL) has been employed as a promising solution for acquiring representations from unlabeled data. Given the greater availability of unlabeled medical images \citep{azizi2022robust}, SSL proves to be an efficient approach for obtaining generalizable representations, which can subsequently be adapted to downstream tasks, even when labeled data is limited. SSL employs a range of pretext tasks to acquire transferable representations without manual annotations. Over recent years, numerous variations of self-supervised learning have surfaced using contrastive learning~\citep{chen2020improved,chen2020simple,grill2020bootstrap,caron2020unsupervised} and restorative learning~\citep{zhou2021models,tang2022self,feng2020parts2whole, xiao2023delving}.

\textbf{Challenges in the Current Literature for Disease Classification.} We study thorax disease classification in this paper. As detailed in Section~\ref{sec:related-works-radiograpic-imaging} about the background for radiographic imaging, the disease areas on radiographic images can be subtle which exhibit localized variations, and such conditions are further complicated by the inevitable noise which is ubiquitously on radiographic images. Effective and robust extraction of features for the disease areas is crucial for disease classification on radiographic images. Although various neural architectures, such as CNNs and ViTs, and different training techniques, such as self-supervised learning with contrastive/restorative learning, have been employed for disease classification on radiographic images, there have been no principled methods which can effectively reduce the adverse effect of noise and background, or non-disease areas, for disease classification on radiographic images.

\begin{figure}[!t]
\centering
\includegraphics[width=1\columnwidth]{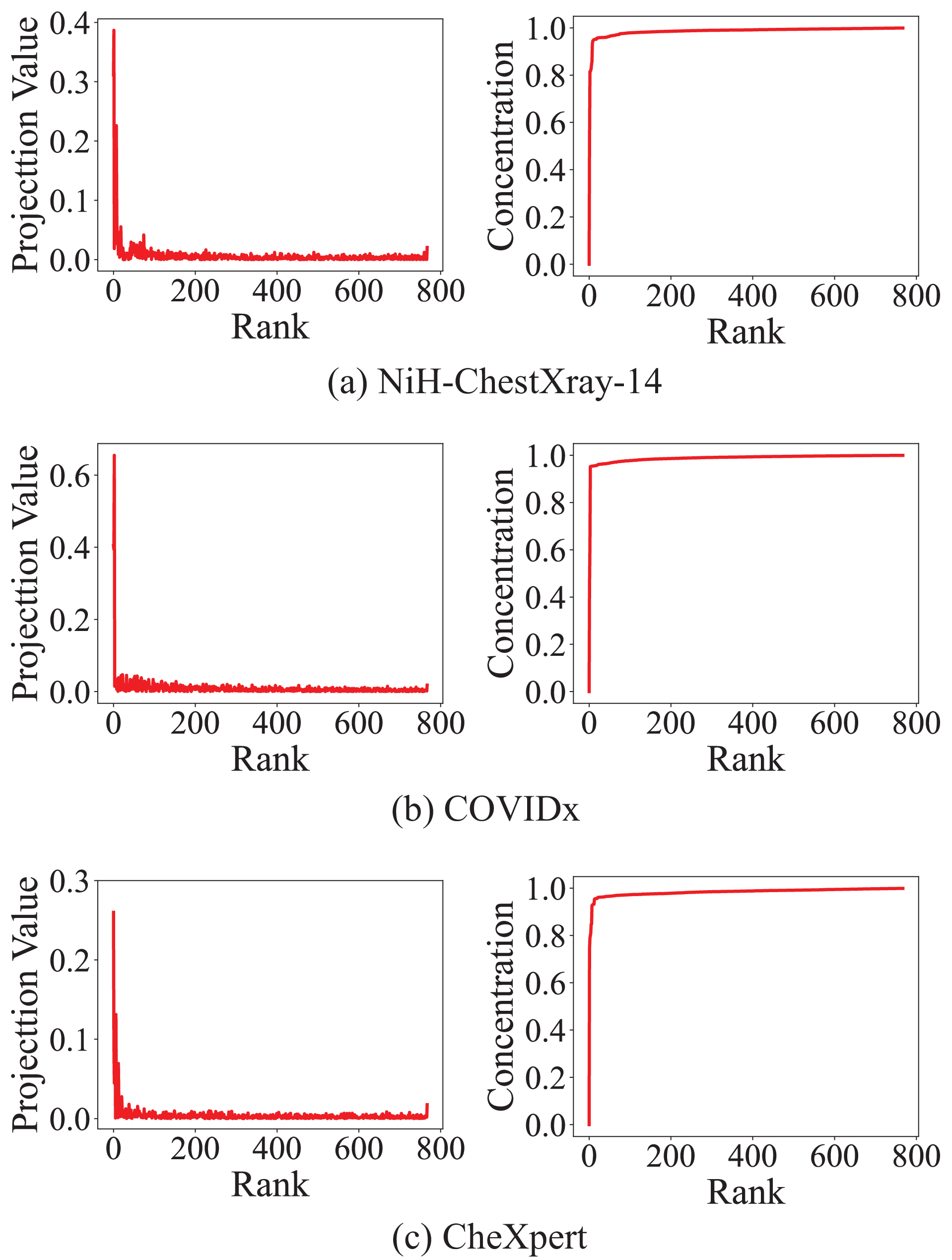}
\caption{Eigen-projection (first column) and signal concentration ratio (second column) of Vit-Base/16 on NiH-ChestXray-14, COVIDx, and CheXpert. To compute the eigen-projection, we first calculate the eigenvectors $\bU$ of the kernel gram matrix $\bK \in \RR^{n \times n}$ computed by a feature matrix $\bF \in \RR^{n \times d}$, then the projection value is computed by $\bp = \frac{1}{C}\sum_{c=1}^{C} \ltwonorm{{\bU}^{\top} \bY^{(c)}}^2/ \ltwonorm{ \bY^{(c)}}^2 \in \RR^n$,  where $C$ is the number of classes, and $\bY\in\{0,1\}^{n \times C}$ is the one-hot labels of all the training data, $\bY^{(c)}$ is the $c$-th column of $\bY$. The eigen-projection $\bp_{r}$ for $r \in [\min(n,d)]$ reflects the amount of the signal projected onto the $r$-th eigenvector of $\bK$, and the signal concentration ratio of a rank $r$ reflects the proportion of signal projected onto the top $r$ eigenvectors of $\bK$. The signal concentration ratio for rank $r$ is computed by $\ltwonorm{\bp^{(1:r)}}$, where $\bp^{(1:r)}$ contains the first $r$ elements of $\bp$. For example, by the rank $r=38$, the signal concentration ratio of $\bY$ on NIH ChestX-ray14, COVIDx, and CheXpert are $0.959$, $0.964$, and $0.962$ respectively.
}
\label{fig:eigen-proj}
\end{figure}

\textbf{Our Contributions.} The contributions of this paper are presented as follows.
First, in order to address the aforementioned challenge, we propose a novel Low-Rank Feature learning (LRFL) method in this paper, which is universally applicable to the training of all neural networks with the application for thorax disease classification. Our LRFL method employs low-rank features for disease classification. The usage of low-rank features are empirically motivated by the low frequency property as shown in Figure~\ref{fig:eigen-proj}. That is, the low-rank projection of the ground truth class labels possesses the majority of the information of the class labels. Inspired by this observation, our LRFL method adds the truncated nuclear norm as a low-rank regularization term to the training loss of a neural network so as to promote low-rank features. Because the actual features used for classification are approximately low-rank and the high-frequency features are significantly truncated, all the noise and the information about the background, or the non-disease areas on radiographic images in the high-frequency features are largely discarded and not learned in a neural network. As a result, the adverse effect of such noise and background is considerably reduced in a network trained by our LRFL method. Furthermore, we propose a new separable approximation to the truncated nuclear norm, so that standard SGD can be used to optimize the training loss with the approximate truncated nuclear norm. Extensive experimental results demonstrate that our LRFL method renders new record mAUC on three standard thorax disease datasets, NIH-ChestX-ray~\citep{wang2017chestx}, COVIDx~\citep{pavlova2022covidx}, and CheXpert~\citep{irvin2019chexpert}, surpassing the current state-of-the-art~\citep{xiao2023delving} with the same pre-training setup.

Second, we provide theoretical result on the sharp generalization bound for our LRFL method, justifying the promising benefit of low-rank learning method in a representation learning framework. Due to our theoretical result and the fact that LRFL can be applied to the training of all neural networks, we expect that our LRFL method can be applied to classification of diseases other than thorax diseases, and generate even broader impact on general classification problems with radiographic images.



\subsection{Notations}
We use bold letters to denote matrices or vectors. $\bth{\cdot}_i$
stands for the $i$-th row of a matrix. $\norm{\cdot}{p}$ denotes the $p$-norm of a vector or a matrix. $\fnorm{\cdot}$ is the Frobenius norm of a matrix.  We use $[m \ldots n]$ to indicate numbers between $m$ and $n$ inclusively, and $[n]$ denotes the natural numbers between $1$ and $n$ inclusively.


\section{Related Works}
\subsection{Radiographic Imaging}
\label{sec:related-works-radiograpic-imaging}
Radiographic imaging~\citep{li2023transforming} has long stood as a cornerstone in medical image analysis. Different from photographic images, where large objects usually lie in the center of images with diverse backgrounds~\citep{deng2009imagenet}, radiography images are generated based on fixed medical imaging protocols~\citep{zhou2021towards,li2022transforming,shamshad2022transformers, xiao2023delving}. As a result, the backgrounds of different radiography images usually exhibit consistent anatomy~\citep{zhou2012mining}. In radiography images, some vital clinical details spread across their expanse. Meanwhile, areas indicating illness, which stand out in the foreground, frequently show more nuanced, detailed, and localized variations~\citep{xiao2023delving, suetens2017fundamentals, zhou2022interpreting}. Such differences make radiographic imaging analysis much more challenging than photographic imaging analysis.

Noise is ubiquitous and inevitable in radiography images generated by medical imaging devices, which stems from various sources, including quantum fluctuations, electronic system interference, scatter radiation, motion blur, and overlapping anatomical structures~\citep{siewerdsen1997empirical, siewerdsen1998signal, manson2019image, chandra2020analysis}. Among these, quantum noise is often identified as the primary source of noise in radiographic imaging~\citep{chandra2020analysis}.
Quantum noise originates from the inherent statistical fluctuations in the number of X-ray photons absorbed or transmitted through the object and detected by the imaging system~\citep{sprawls1993physical, shung2012principles,suetens2017fundamentals,  chandra2020analysis}. These variations introduce graininess or mottling to the image, potentially obscuring fine details and diminishing tissue contrast. The level of quantum noise on a radiography image is affected by factors such as the X-ray dose, the detector sensitivity, and the thickness of the imaged object~\citep{sprawls1993physical}. As the quantum noise is a result of the random manner in which X-ray photons are emitted and absorbed, its presence can be modeled as a Poisson process~\citep{suetens2017fundamentals, chandra2020analysis}.  In addition, under conditions of high photon flux, the behavior of quantum noise can be approximated by a Gaussian distribution~\citep{lee2018poisson, ding2018statistical}. This approximation allows for the use of various image processing techniques, such as filtering~\citep{ding2018statistical}, aimed at reducing noise and improving image quality.


\subsection{Medical Image Analysis with Deep Learning}
Following the remarkable achievements of deep learning in photographic image processing~\citep{he2016deep, Lin2017a, lin2017feature}, there has been a growing interest in harnessing deep neural networks to enhance medical image analysis due to their ability to learn complex representations. Ever since U-Net~\citep{ronneberger2015u, falk2018u, zhou2018unet++} wihch first shows the power of Convolutional Neural Networks (CNNs) in medical imaging, methods based on CNNs have demonstrated dominated performance in almost every field of medical imaging, including image classification~\citep{shen2018dynamic,wang2019thorax,ma2020multilabel}, object detection~\citep{falk2019u, zhou2018unet++, yang2021artificial}, and semantic segmentation~\citep{yang2021artificial, yao2021unsupervised, zhou2018unet++, simpson2019large, sourati2019intelligent}.  In addition, methods based on other techniques such as Recurrent Neural Networks (RNNs)~\citep{zhou2019handbook, gao2019distanced} and Reinforcement Learning (RL)~\citep{ zhou2021deep, xu2022deep, hu2023reinforcement} have also been developed for medical imaging.

More recently, following the success of Transformer in natural language processing~\citep{vaswani2017attention}, visual transformers have demonstrated remarkable performance compared to state-of-the-art CNNs across a wide range of computer vision tasks, including image classification~\citep{yuan2021tokens, dosovitskiy2020image}, object detection~\citep{liu2021swin, zhu2020deformable}, and semantic segmentation~\citep{cai2022efficientvit}.
Despite the debate over the adoption of Transformers and CNNs in visual domains, regarding generalization ability~\citep{liu2022convnet,zhou2021ibot,bao2021beit,xiao2022transforming,touvron2021training,ding2022scaling, bai2021transformers,mao2022towards,zhang2022delving,zhou2022understanding}, training data requirements~\citep{dosovitskiy2020image,steiner2021train,tay2022scaling}, computational costs~\citep{paul2022vision}, visual transformers have shown the potential to achieve even better performance than CNNs in medical imaging analysis~\citep{xiao2023delving, chen2021pre, chen2021transunet}. For example, the TransUNet~\citep{chen2021transunet} follows the 2D UNet~\citep{falk2019u} design and incorporates the Transformer blocks in building the encoder and decoder, which achieves promising results in CT segmentation tasks.
Given that most state-of-the-art CNNs use small-sized convolution kernels, such locality predisposes CNNs towards local spatial patterns.
In contrast, the global self-attention mechanism in visual transformers greatly boosts their ability to model the long-range dependencies in medical images~\citep{li2023transforming}.

Self-supervised contrastive learning~\citep{chen2020improved,chen2020simple,grill2020bootstrap,caron2020unsupervised, xiao2023delving} has demonstrated significant promise in dealing with the scarcity of high-quality annotations in the medical imaging domain~\citep{zhou2021towards, xiao2023delving, chen2021pre}.
Contrastive learning approaches treat each image as a separate class, aiming to reduce the similarity between augmented views of distinct images while increasing the similarity between views of the same image during neural network pre-training. However, since radiography images are obtained following standard radiography imaging protocols~\citep{xiang2021painting,haghighi2022dira}, different images show much higher similarity compared to photographic images~\citep{he2020momentum,chen2020improved}. To avoid such problems in pre-training neural networks for medical imaging, recent works adopt restorative strategies~\citep{alex2017semisupervised,chen2019self,zhou2019models,zhu2020rubik,chen2020imagegpt,xie2021simmim, xiao2023delving}, which perform pixel-wise image reconstruction. For instance,~\citep{xiao2023delving} adopts masked autoencoders (MAE)~\citep{he2022masked} to pre-train both CNNs and ViTs and achieve state-of-the-art performance in radiography image classification. In our work, we also adopt the MAE method in~\citep{xiao2023delving} to pre-train our neural networks before learning low-rank features.



\section{Formulation}
\subsection{Pipeline for Thorax Disease Classification}
\begin{figure}[!htbp]
\begin{center}
\resizebox{1\columnwidth}{!}{\includegraphics[width=1\textwidth]{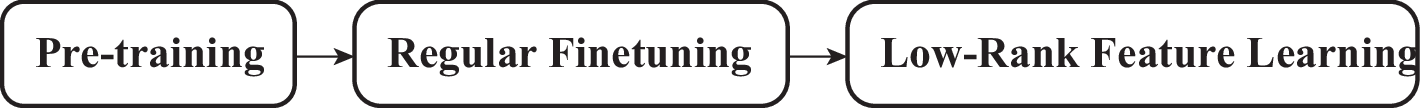}
}
\caption{Training Pipeline for Thorax Disease Classification.}
\label{fig:pipeline}
\end{center}
\end{figure}

We follow~\citep{xiao2023delving} and use masked MAE~\citep{he2022masked} to pre-train CNNs or ViTs, and then perform our novel LRFL. The full training pipeline of learning low-rank features for disease classification can be described in three steps. In the first step, we pre-train the networks with the self-supervised restorative learning method masked MAE~\citep{he2022masked} on the pre-training dataset such as ImageNet-1k~\citep{krizhevsky2012imagenet} and X-rays (0.5M)~\citep{xiao2023delving}. We randomly mask patches on input images and optimize the networks for pixel-wise image reconstruction on the masked patches. In the second step, we finetune the pre-trained networks with the cross-entropy loss for image classification on the target datasets such as NIH-ChestX-ray~\citep{wang2017chestx}, COVIDx~\citep{pavlova2022covidx}, and CheXpert~\citep{irvin2019chexpert}. In the last step, we fix the backbones of the network and finetune the linear classifier with our novel LRFL method, as illustrated in Figure~\ref{fig:pipeline}.

\subsection{Problem Setup for LRFL}
We now introduce the problem setup for LRFL with training details.
Suppose the training data are given as $\set{\bx_i,\by_i}_{i=1}^n$ where $\bx_i$ and $\by_i \in \RR^C$ are the $i$-th training data point and its corresponding class label vector respectively, and $C$ is the number of classes. Each element  $\by_i$ is binary with $\by_i = 1$ indicating the $i$-th disease is present in $\bx_i$, otherwise $\by_i = 0$. Given the feature $\bF \in \RR^{n \times d}$ of all the training data where $d$ is the dimension of the feature and $\bF$ is the feature produced by the neural network obtained by the regular finetuning as step two of the pipeline in Figure~\ref{fig:pipeline}, we can train a linear neural network as a linear classifier by optimizing
\bal\label{eq:network-obj}
\min\limits_{\bWlin \in \RR^{d \times C}} L(\bW) =  \frac{1}{n} \sum_{i=1}^n \KL \pth{\by_i, \bth{\sigma\pth{\bF \bWlin}}_i}.
\eal
Here $\bth{\sigma(\bA)}_{ic} = 1/(1+\exp(-\bA_{ic}))$ is the element-wise sigmod function for $\bA \in \RR^{n \times C}$, $i \in [n], c \in [C]$. When $\sigma$ is applied on a matrix $\bA$, it returns a matrix of the same size as $\bA$ and where the sigmod function is applied on every element of $\bA$. $\bth{\cdot}_i$
stands for the $i$-th row of a matrix. $\KL$ stands for the element-wise binary cross-entropy function. Given two nonnegative vectors $\bu = \bth{u_1,\ldots,
u_d} \in \RR^d,\bv = \bth{v_1,\ldots,v_d} \in \RR^d$ where $u_i \in \set{0,1}$
for all $i \in [d]$ and $\supnorm{\bv} \le 1$,
$\KL(\bu,\bv) \defeq \sum_{j = 1}^d -u_i \log v_i - (1-u_i) \log (1-v_i)$. We use
$\bY = \bth{\by_1; \by_2; \ldots ; \by_n} \in \RR^{n \times C}$ to denote the training label matrix by stacking the label vectors of all the training data.



\subsection{Generalization Bound for Low-Rank Feature Learning}
\label{sec:generalization-bound}
We define the loss function $\ell(\Net(\bx),\by) \defeq \ltwonorm{\Net(\bx) - \by}^2$, and the generalization error of the network $\Net$ is the expected risk of the loss $\ell$, which is denoted by $L_{\cD}(\Net_{\bW})
\defeq \Expect{(\bx,\by) \sim \cD}{\ell(\Net_{\bW}(\bx),\by)}$. The kernel gram matrix for the feature $\bF$ is $\bK_n = \frac 1n \bF \bF^{\top}$. We let $\hat \lambda_1 \ge \hat \lambda_2 \ge \ldots \ge \hat \lambda_r > 0$ where $\bar r \le \min\set{n,d}$ is the rank of $\bK_n$. 
Suppose the Singular Value Decomposition of $\bF$ is $\bF = \bU \bSigma \bV^{\top}$, where $\bU \in \RR^{n \times d}$ has orthogonal columns, $\bSigma \in \RR^{d \times d}$ is a diagonal matrix with diagonal elements being the singular values of $\bF$, and $\bV \in \RR^{d \times d}$ is an orthogonal matrix. The columns of $\bU$ and $\bV$ are also called the left eigenvectors and the right eigenvectors of $\bF$ respectively. Let $\sigma_1 \ge \sigma_2 \ldots \ge \sigma_d$ be the singular values of $\bF$,
and $\bar \bY = \bU^{\bar r} {\bU^{\bar r}}^{\top} \bY$ be the projection of the training label matrix $\bY$ onto the subspace spanned by the top-$\bar r$ left eigenvectors of $\bF$, where
$\bU^{\bar r} \in \RR^{n \times {\bar r}}$ is formed by the top $\bar r$ eigenvectors in $\bU$. Then we have the following theorem giving the sharp generalization error bound for the linear neural network in (\ref{eq:network-obj}).
\begin{theorem}
\label{theorem:optimization-linear-kernel}
For every $x > 0$, with probability at least $1-\exp(-x)$, after the $t$-th iteration of gradient descent for all $t \ge 1$, we have
\bal\label{eq:optimization-linear-kernel-generalization}
L_{\cD}(\Net_{\bW}) &\le \fnorm{\bY - \bar \bY} + c_1 \pth{1-\eta \hat\lambda_r}^{2t} \fnorm{\bY}^2
\nonumber \\
&+ c_2 \min\limits_{h \in [0,r]} \pth{\frac hn + \sqrt{\frac 1n
\sum\limits_{i=h+1}^r \hat \lambda_i}} + \frac{c_3 x}{n},
\eal
where $c_1,c_2,c_3$ are positive constants.
\end{theorem}
\begin{remark}\label{remark:optimization-linear-kernel}
The RHS of (\ref{eq:optimization-linear-kernel-generalization}) is the generalization error bound for the linear neural network used in LRFL as step three of the pipeline in Figure~\ref{fig:pipeline}. Moreover, let $\sigma_1 \ge \sigma_2 \ldots \ge \sigma_d$ be the singular values of $\bF$. Due to the fact that
$\sqrt{\frac 1n \sum\limits_{i=h+1}^r \hat \lambda_i} \le \frac 1n \sum\limits_{i=h+1}^r \sigma_i$, it follows by (\ref{eq:optimization-linear-kernel-generalization}) that
\bal\label{eq:optimization-linear-kernel-generalization-sigma}
L_{\cD}(\Net_{\bW}) &\le c_1 \pth{1-\eta \hat\lambda_r}^{2t} \fnorm{\bY}^2
\nonumber \\
&+ c_2 \pth{\frac hn + \frac 1n \sum\limits_{i=T+1}^d \sigma_i} + \frac{c_3 x}{n},
\eal
which holds for all $T \in [0,d]$. (\ref{eq:optimization-linear-kernel-generalization-sigma}) motivates the reduction of the truncated nuclear norm of the feature $\bF$, as detailed in the next subsection.
\end{remark}

\subsection{Optimization of the Truncated Nuclear Norm in SGD }
Using notations in Section~\ref{sec:generalization-bound}, the truncated nuclear norm of $\bF$ is $\norm{\bF}{T} \defeq \sum\limits_{i=T+1}^d \sigma_i$ where $T \in [0,d]$. It can be observed by the generalization error bound (\ref{eq:optimization-linear-kernel-generalization-sigma}) of Remark~\ref{remark:optimization-linear-kernel} that a smaller $\norm{\bF}{T}$ renders a tighter upper bound for the generalization error of the linear neural network used for LRFL. This observation gives a strong theoretical motivation for us to add the truncated nuclear norm $\norm{\bF}{T}$ to the training loss (\ref{eq:network-obj}). However, $\norm{\bF}{T}$ is not separable, so the training loss with $\norm{\bF}{T}$ cannot be directly optimized by the standard SGD. To address this problem, we propose an approximation $\overline{\norm{\bK}{T}}$ to $\norm{\bK}{T}$ which is separable so that $\overline{\norm{\bK}{T}}$  can be optimized by standard SGD.

First, we note that if $\bU,\bV$ are known, then $\bSigma = \bU^{\top}\bF \bV$. If we have an approximation $\overline \bU$  to $\bU$ and an approximation $\overline \bV$ to $\bV$, then $\bSigma$ can be approximated by
\bals
\overline \bSigma =  {\overline \bU}^{\top}\bF \overline{\bV}.
\eals
As a result, the approximation $\overline{\norm{\bK}{T}}$ to the truncated nuclear norm is
\bal\label{eq:approximate-TN}
\overline{\norm{\bK}{T}} &= \sum\limits_{s=T+1}^{d} \overline \bSigma_{ss}
= \sum\limits_{s=T+1}^{d} \bth{{\overline \bU}^{\top}\bF \overline{\bV}}_{ss}
\nonumber \\
&=\sum\limits_{i=1}^n \pth{ \sum\limits_{s=T+1}^{d} \sum\limits_{k=1}^d {\overline \bU}^{\top}_{si} \bF_{ik} \overline{\bV}_{ks}  }.
\eal

Due to the above discussions, the loss function of LRFL with the approximate truncated nuclear norm $\overline{\norm{\bK}{T}} $ is
\bal\label{eq:obj-lr-overall}
\cL(\bW) &=  \frac{1}{m} \sum_{v_i \in \mathcal{V_L}} \KL \pth{\by_i, \bth{\sigma\pth{\bF \bWlin}}_i} \nonumber \\
&+ \eta \sum\limits_{i=1}^n \pth{ \sum\limits_{s=T+1}^{d} \sum\limits_{k=1}^d {\overline \bU}^{\top}_{si} \bF_{ik} \overline{\bV}_{ks}  },
\eal
where $\eta > 0$ is the weighting parameter for the truncated nuclear norm. Because (\ref{eq:obj-lr-overall}) is to be optimized by the standard SGD, we have the loss function of LRFL
for the $j$-th minibatch $\cB_j \subseteq [n]$ as follows:
\bal\label{eq:loss-Bj}
\cL_j(\bW) &= \frac{1}{\abth{\cB_j}} \sum_{i \in \cB_j} \KL \pth{\by_i, \bth{\sigma\pth{\bF \bWlin}}_i} \nonumber \\
&+ \frac{\eta}{\abth{\cB_j}}  \sum\limits_{i \in \cB_j} \pth{ \sum\limits_{s=T+1}^{d} \sum\limits_{k=1}^d {\overline \bU}^{\top}_{si} \bF_{ik} \overline{\bV}_{ks}  }.
\eal
The approximation $\overline \bU$ and $\overline \bV$ can be computed as the left and right eigenvectors of the feature $\bF$ computed at earlier epochs. In order to save computation and avoiding performing SVD for $\bF$ at every epoch, we propose to update $\overline \bU$ and $\overline \bV$ only after certain epochs. Algorithm~\ref{algorithm:Separable-TN} describes the training algorithm for our LRFL that uses standard SGD to optimize the loss function (\ref{eq:obj-lr-overall}). Before the first epoch, we compute $\overline \bU$ and $\overline \bV$ as the left and right eigenvectors of the feature $\bF$ at the initialization of the neural network. After every $t_0$ epochs for $t_0$ being a constant integer, we update $\overline \bU$ and $\overline \bV$ as the left and right eigenvectors of the feature $\bF$ of the neural network righter after $t_0$-th epoch. The algorithm is described in Algorithm~\ref{algorithm:Separable-TN}.

\begin{algorithm}[!htb]
\caption{Training Algorithm with the Approximate Truncated Nuclear Norm by SGD}\label{algorithm:Separable-TN}
{
\small
\begin{algorithmic}[1]
\STATE Initialize the weights of the network
by $\cW = \cW(0)$ through random initialization
\STATE Compute feature $\bF$ by the neural network, and its SVD as $\bF = \bU \bSigma \bV$
\STATE Update $\overline{\bU} = \bU$, $\overline{\bV} = \bV$
\FOR{$t = 1,2,\ldots, t_{\max}$}
\IF{$t\equiv 0 \pmod{t_0}$}
\STATE Compute feature $\bF$ of the neural network, and its SVD $\bF = \bU \bSigma \bV$.
\STATE Update $\overline{\bU} = \bU$, $\overline{\bV} = \bV$
\ENDIF
\FOR{$b = 1,2,\ldots, B$}
\STATE Update $\cW$ by applying gradient descent on batch $\cB_j$ using the gradient of the loss $\cL_j$ in Eq.(\ref{eq:loss-Bj})
\ENDFOR
\ENDFOR
\STATE \textbf{return} The trained weights $\cW$ of the network
\end{algorithmic}
}
\end{algorithm}

\section{Experimental Results}

In this section, we conduct experiments on medical datasets to show the effectiveness of the proposed LRFL. The experiments section is organized as follows. In Section \ref{subsection:implementation_details}, we discuss our experimental setup and implementation details. In Section \ref{sec:chestxray14_result}, Section \ref{sec:covidx_result} and Section~\ref{sec:chexpert_result}, we compare LRFL against various architectures on three medical datasets. In Section \ref{subsection:low_data_regime}, we investigate the performance of LRFL on small data regimes. In Section \ref{subsection:ablation_study}, we empirically show the effectiveness of our loss function and investigate the significance of the low-rank regularization.

\subsection{Implementation Details}
\label{subsection:implementation_details}

In this section, we evaluate the proposed LRFL for thorax disease classification. We utilize networks pre-trained on ImageNet~\citep{russakovsky2015imagenet} or chest X-rays in~\citep{xiao2023delving} with MAE, which adopts the self-supervised learning strategy by reconstructing missing pixels from patches of input images.
We fine-tune the pre-trained networks with low-rank regularization for classification on three public X-ray datasets, namely (1) NIH-Chest Xray 14 \citep{wang2017chestx}, (2) Stanford CheXpert \citep{irvin2019chexpert} and (3) COVIDx \citep{pavlova2022covidx}. We use ADAM optimizer in the fine-tuning process. The batch size is set to $2048$ for all datasets.
We first fine-tune the entire networks for 75 epochs with Adam following the settings in~\citep{xiao2023delving}. Next, we fine-tune the network with low-rank regularization for another 75 epochs.
The learning rate is initialized as $2.5 \times 10^{-5}$ and annealed down to $1 \times 10^{-7}$ following a cosine schedule. The default values of momentum and weight decay are set to $0.9$ and $0$. We use standard data augmentation techniques for both datasets, including random-resize cropping, random rotation, and random horizontal flipping.
Throughout the paper, we evaluate our LRFL method on both CNN and visual transformer architectures including ResNet-50, DenseNet, ViT-S, and ViT-B. We refer to our model by `X-LR', where X is the base model. For example, a ResNet-50 model trained using low-rank features is referred to as ResNet-50-LR.

\textbf{Tuning the $T$ and $\eta$ by Cross-Validation.} We tune the optimal values of feature rank $T$ and weighting parameter for the truncated nuclear
norm $\eta$ on each dataset. Let $ T=\lceil \gamma \min(n,d)\rceil$, where $\gamma$ is the rank ratio. We select the values of $\gamma$ and  $\eta$ by performing 5-fold cross-validation on $20\%$ of the training data in each dataset. The value of $\gamma$ is selected from $\{0.01, 0.02, 0.03, 0.04, 0.05, 0.1,0.15,0.2\}$.  The value of $\eta$ is selected from $\{5\times 10^{-4}, 1\times 10^{-3}, 2.5\times 10^{-3}, 5\times 10^{-3}, 1\times 10^{-2}\}$. The optimal values of $\eta$ and $\gamma$ selected by cross-validation on each dataset are shown in Table~\ref{tab:hyper-parameters}.
\begin{table}[!htbp]
\center
\begin{center}
\resizebox{1\linewidth}{!}{
\begin{tabular}{|c|ccc|}
\hline
Parameters    & NIH-ChestX-ray & COVIDx  & CheXpert  \\ \hline
$\gamma$      & $0.05$ & $0.003$    & $0.05$   \\
$\eta$       & $5\times 10^{-4}$ & $1\times 10^{-3}$      & $1\times 10^{-3}$    \\ \hline
\end{tabular}
}
\caption{Selected rank ratio $\gamma$ and weighting parameter for the truncated nuclear norm $\eta$ on each dataset.}
\label{tab:hyper-parameters}
\end{center}
\end{table}

\subsection{NIH ChestX-ray14}
\label{sec:chestxray14_result}

\begin{figure}[!t]
\begin{center}
\resizebox{1\columnwidth}{!}{\includegraphics[width=1\textwidth]{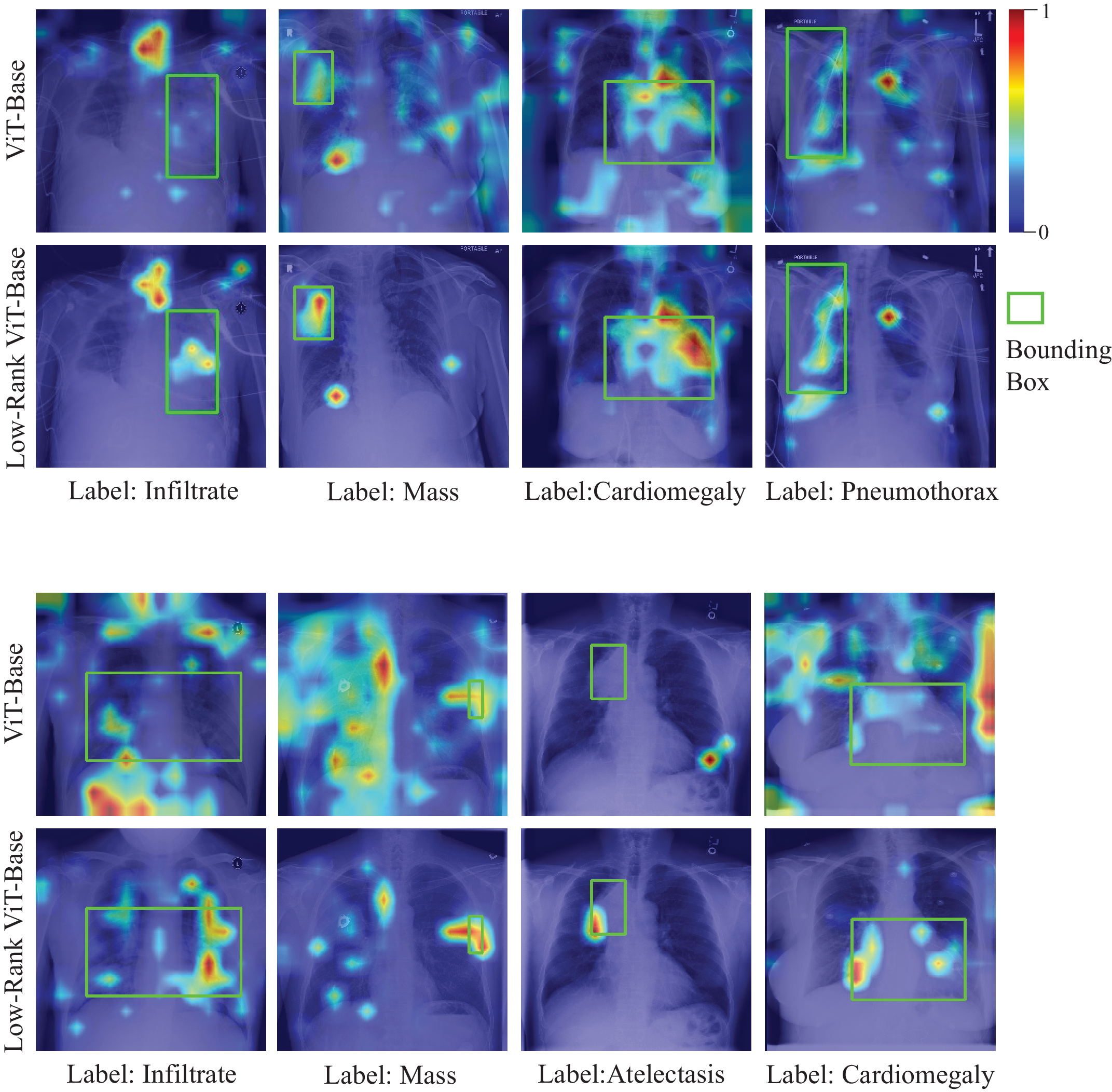}
}
\caption{Grad-CAM visualization results on NIH ChestX-ray 14. The figures in the first row are the visualization results of ViT-Base, and the figures in the second row are the visualization results of Low-Rank ViT-Base. Ground-truth bounding box for each disease is shown in green. Although both the base model and its corresponding low-rank model predict the correct disease label, the low-rank model pays more attention to the disease location than the base model. More Grad-CAM visualization results are deferred to Figure~\ref{fig:sup_grad-cam} of the supplementary.}
\label{fig:grad-cam}
\end{center}
\end{figure}

\begin{table}[!t]
\center
\begin{center}
\resizebox{1\linewidth}{!}{
\begin{tabular}{|c|c|c|c|c|}
\hline
    Pre-training Dataset & Model & Rank  & mAUC   \\ \hline
     \multirow{2}{*}{Imagenet-1k} & ResNet-50   & - & 81.78 \\ \cline{2-4}
      & ResNet-50-LR  & 0.05r & 82.18 \\\hline
     \multirow{2}{*}{Imagenet-1k} & DenseNet-121  & - & 82.02 \\ \cline{2-4}
      & DenseNet-121-LR  & 0.05r & 82.35 \\ \hline
    \multirow{2}{*}{X-rays(0.3M)} & ViT-S \citep{xiao2023delving}  & - & 82.30 \\ \cline{2-4}
      &  ViT-S-LR & 0.05r & 82.70 \\ \hline
     \multirow{2}{*}{X-rays(0.5M)} & ViT-B \citep{xiao2023delving}  & - & \underline{83.00} \\ \cline{2-4}
      &  ViT-B-LR & 0.05r & \textbf{83.40} \\ \hline
\end{tabular}
}
\caption{The table compares the performance of various models and their low-rank counterparts on NIH ChestX-ray14 dataset. The best result is presented in bold, and the second-best result is underlined. This convention applies to all the tables in our paper.}
\label{table:nih_lowrank}
\end{center}
\end{table}

\noindent\textbf{Experimental setup.}
NIH-ChestX-ray14 \citep{wang2017chestx} consists of $112,120$ X-rays collected from $30,805$ unique patients. Each X-ray has up to $14$ associated labels, with the possibility of multiple labels per image. Following the official data split in~\citep{wang2017chestx}, we use $75,312$ images for training and $25,596$ images for testing. Raw images from the dataset are in the size of $1024 \times 1024$. In our experiments, we scale down the input images to the size of $224 \times 224$. We report the mean AUC (Area Under the Curve) for $14$ distinct classes and conduct a comprehensive comparison with 18 widely recognized and influential baseline methods.

\begin{table}[!htbp]
\centering

\resizebox{1\linewidth}{!}{
\begin{tabular}{|c|c|c|c|c|}
\hline
    Method & Architecture & Pre-training & mAUC \\
    \hline
    Wang et al. \citep{wang2017chestx} & RN50 & \multirow{17}{*}{ImageNet-1K} & 74.5 \\
    Li et al.\citep{li2018thoracic} & RN50 & & 75.5 \\
    Yao et al. \citep{yao2018weakly} & RN\&DN & & 76.1 \\
    Wang et al.\citep{wang2019thorax} & R152 & & 78.8 \\
    Ma et al.\citep{ma2019multi} & R101 & & 79.4 \\
    Tang et al.\citep{tang2018attention} & RN50 & & 80.3 \\
    Baltruschat et al.\citep{baltruschat2019comparison} & RN50 & & 80.6 \\
    Guendel et al.\citep{guendel2018learning} & DN121 & & 80.7 \\
    Guan et al.\citep{guan2018multi} & DN121 & & 81.6 \\
    Seyyed et al.\citep{seyyedkalantari2020chexclusion} & DN121 & & 81.2 \\
    Ma et al.\citep{ma2020multilabel} & DN121$(\times 2)$ & & 81.7 \\
    Hermoza et al.\citep{hermoza2020region} & DN121 & & 82.1 \\
    Kim et al.\citep{Kim_2021_CVPR} & DN121 & & 82.2 \\
    Haghighi et al.\citep{haghighi2022dira} & DN121 & & 81.7 \\
    Liu et al.\citep{liu2022acpl} & DN121 & & 81.8 \\
    Taslimi et al.\citep{taslimi2022swinchex} & SwinT & & 81.0 \\
    \hline
    MoCo v2 \citep{xiao2023delving} & DN121 & & 80.6 \\

    MAE \citep{xiao2023delving} & DN121 & X-rays (0.3M) & 81.2 \\

    MAE \citep{xiao2023delving} & ViT-S/16 & & 82.3 \\
    \hline
    MAE \citep{xiao2023delving} & ViT-B/16 & X-rays (0.5M) & \underline{83.0} \\
    \hline
    RN-50-LR (Ours) & RN50 & \multirow{2}{*}{ImageNet-1K} & 82.2 \\

    DN-121-LR (Ours) & DN121 & & 82.4 \\
\hline
    ViT-S-LR (Ours) & ViT-S/16 & X-rays (0.3M) & 82.7 \\

    ViT-B-LR (Ours) & ViT-B/16 & X-rays (0.5M) & \textbf{83.4} \\
\hline
\end{tabular}
}

\caption{The table shows the performance of various state-of-the-art (SOTA) CNN-based and Transformer-based methods on ChestX-ray14. With the same pre-training settings as ViT-B in \citep{xiao2023delving}, our ViT-B-LR achieves the new record high of 83.4 mAUC. RN, DN, and SwinT represent ResNet, DenseNet, and Swin Transformer.
}
\label{table:nih_sota}
\end{table}

\smallskip\noindent\textbf{Results and analysis.} Table \ref{table:nih_lowrank} presents the performance comparisons between several top-performing baseline models and their corresponding low-rank models on the NIH ChestX-ray14 dataset. Throughout this section we use postfix ``-LR'' to indicate a neural network trained with our LRFL.
For example, we use ViT-B model pre-trained on $266,340$ chest X-rays with Masked Autoencoders (MAE) \citep{xiao2023delving}. The pre-trained ViT-B network is fine-tuned on the NIH ChestX-ray14 dataset and achieves a mean AUC of $83.0$. Next, we fine-tune ViT-B with low-rank regularization for another 75 epochs. The low-rank model, denoted as ViT-B-LR, achieves a mean AUC of $83.4$. It is observed that all low-rank models achieve improvement in mean AUC compared to the corresponding base models. ViT-S-LR improves its base model by a mean AUC of $0.4\%$. Similar improvements are observed for CNN-based models as well. For example, ResNet-50-LR improves its base model by a mean AUC of $0.40\%$.

Table \ref{table:nih_sota} shows the performance comparisons of models trained with LRFL against state-of-the-art CNN and Transformer models on NIH ChestX-ray14.
In our experiments, ViT-B-LR achieves the new state-of-the-art performance with a mean AUC of 83.4\%. It is important to highlight that the research community dedicated four years to enhancing the AUC score for CNN-type architectures, advancing it from 74.5 to 82.2. This improvement was primarily attributed to the challenging nature of the training process.

To study how LRFL improves the performance of base models in disease detection, we use the Grad-CAM~\citep{selvaraju2017grad} to visualize the parts in the input images that are responsible for the predictions of the base models and low-rank models. Examples of visualization results in Figure~\ref{fig:grad-cam} show that our LRFL models usually focus more on the areas inside the bounding box associated with the labeled disease. In contrast, the base models also focus on the areas outside the bounding box or even areas in the background. More Grad-CAM visualization results are deferred to Figure~\ref{fig:sup_grad-cam} of the supplementary.
\subsection{COVIDx}
\label{sec:covidx_result}




\noindent\textbf{Experimental setup.}
COVIDx (Version 9A) \citep{pavlova2022covidx} consists of 30,386 chest X-rays collected from 17,026 unique patients. We follow the previous works\citep{pavlova2022covidx, xiao2023delving} in splitting the dataset into 29,986 training images with four different classes and 400 testing images with three classes. For fair comparisons with the previous methods, we report Top-1 accuracy on the test set (3 classes).

\begin{table*}[!htbp]
\center
\small
\begin{center}
\resizebox{0.85\linewidth}{!}{
\begin{tabular}{|c|c|c|c|c|}
    \hline
    Method & Architecture & Rank & Accuracy & Covid-19 Senstitvity \\
        \hline COVIDNet-CXR Small \citep{Wang2020covid} & - & - & 92.6 & 87.1 \\
        COVIDNet-CXR Large \citep{Wang2020covid} & - & - & 94.4 & 96.8 \\
        DN121 (MoCo v2) \citep{xiao2023delving} & DN121 & - & 94.0 & 94.5 \\
        DN121 \citep{xiao2023delving} & DN121 & - & 93.5 & 97.0 \\ \hline
        ViT-S \citep{xiao2023delving} & ViT-S/16 & - & 95.2 & 94.5 \\
        ViT-S-LR (Ours) & ViT-S/16 & 0.01r & \underline{96.8} & \underline{97.5} \\ \hline
        ViT-B \citep{xiao2023delving} & ViT-B/16 & - & 95.3 & 95.5 \\
        ViT-B-LR (Ours) & ViT-B/16 & 0.003r & \textbf{97.0} & \textbf{98.5} \\ \hline
\end{tabular}
}
\caption{The table shows the performance of various state-of-the-art (SOTA) CNN-based and Transformer-
based methods on COVIDx. With the same pre-training settings as ViT-S and ViT-B in \citep{xiao2023delving}, our ViT-S-LR and ViT-B-LR achieve 97.0 \% Accuracy. DN represents DenseNet.}
\label{table:sota_covidx}

\end{center}
\end{table*}
\begin{table*}[!htbp]
\center
\begin{center}
\resizebox{0.85\linewidth}{!}{
    \begin{tabular}{|c|c|c|c|c|c|c|c|}
    \hline
        Method  & Architecture & Rank & Atelectasis & Cardiomegaly  & mAUC (\%)\\ \hline

        Allaouzi et al.\citep{allaouzi2019novel} & \multirow{10}{*}{DN121} & - & 72.0 & 88.0 & 82.8 \\

        Irvin et al.\citep{irvin2019chexpert} &  & - & 81.8 & 82.8 & 88.9 \\

        Seyyedkalantari et al.\citep{seyyedkalantari2020chexclusion} &  & - & 81.2 & 83.0  & 87.3 \\

        Pham et al.\citep{pham2021interpreting}  &  & - & 82.5 & 85.5  & 89.4 \\

        Hosseinzadeh et al.\citep{hosseinzadeh2021systematic} &  & - & - & -  & 87.1 \\

        Haghighi et al.\citep{haghighi2022dira} &  & - & - & -  & 87.6 \\

        Kang et al.\citep{kang2021data} &  & - & 82.1 & 85.9  & 89.0 \\
        DN121 (MoCo v2) \citep{xiao2023delving} & & - & 78.5 & 77.9  & 88.7\\
        DN121 \citep{xiao2023delving} & & - & 81.5 & 77.6 & 88.7\\ \hline
        ViT-S \citep{xiao2023delving} & ViT-S/16 & - & 83.5 & 81.8  & 89.2 \\
        ViT-S-LR (Ours) & ViT-S/16 & 0.05r & \textbf{86.3} & \underline{93.7} & \underline{89.6} \\ \hline
         ViT-B \citep{xiao2023delving} & ViT-B/16 & - & 82.7 & 83.5 & 89.3 \\
         ViT-B-LR (Ours) & ViT-B/16 & 0.05r & \underline{85.4} & \textbf{94.6} & \textbf{89.8} \\
        \hline
    \end{tabular}
}
    \caption{The table shows the performance of various
state-of-the-art (SOTA) CNN-based and Transformer-
based methods on CheXpert.}
    \label{tab:chexpert}
\end{center}
\end{table*}

\smallskip\noindent\textbf{Results and analysis.} Table \ref{table:sota_covidx} compares the performance of SOTA transformer-based models and the LRFL models on the COVIDx dataset. Similar to Section \ref{sec:chestxray14_result}, the base ViTs are first pre-trained on 266,340 chest X-rays using Masked Autoencoders (MAE) and next the pre-trained model is fine-tuned on COVIDx dataset. It can be observed by Table \ref{table:sota_covidx} that both ViT-S-LR and ViT-B-LR outperform their corresponding base models ViT-S and ViT-B, achieving an increase in accuracy of 1.6\% and 1.7\%, respectively.
Table \ref{table:sota_covidx} also compares the performance of our LRFL models against the state-of-the-art models on the COVIDx dataset. LRFL models achieve much higher accuracy as compared to CNN-based models such as DenseNet-121. ViT-B-LR achieves the new SOTA performance of 97\% top-1 accuracy with input resolution set to 224$\times$224, which exceeds the previous SOTA performance \citep{xiao2023delving} by 1.7 \% in top-1 accuracy.

\subsection{Stanford CheXpert}
\label{sec:chexpert_result}

\noindent\textbf{Experimental setup.} CheXpert~\citep{irvin2019chexpert} consists of 224,316 chest X-rays collected from 65,240 patients, where 191,028 chest X-rays are used for training. Each X-ray in the dataset has radiology reports indicating the presence of 14 diseases. Following the protocol in \citep{xiao2023delving}, all images are resized into $224\times224$. We also report the mean AUC (Area Under the Curve) for the 5 distinct classes and conduct a comprehensive comparison with 9 widely recognized and influential baseline methods.

\noindent\textbf{Results and analysis.}
Table~\ref{tab:chexpert} presents the performance comparisons between the baseline models and the LRFL models on the CheXpert dataset. It is observed that ViT-B-LR achieves state-of-the-art performance of 89.8\% in mAUC, and improves the performance of ViT-B by 0.5\% in mAUC. ViT-S-LR also improves the performance of ViT-S by 0.4\% in mAUC, which demonstrates the power of LRFL. We also show the classification accuracy on Atelectasis and Cardiomegaly in Table~\ref{tab:chexpert}, where our method exhibits much better performance than baseline methods. For example, ViT-S-LR achieves an mAUC of 86.3\% on Atelectasis, with a 2.8\% improvement over ViT-S trained with MAE. Such improvements demonstrate the power of LRFL in detecting distinct diseases.
\begin{figure*}[!htbp]
    \subfigure[ChestX-ray14]{
        \includegraphics[width=0.3\textwidth]{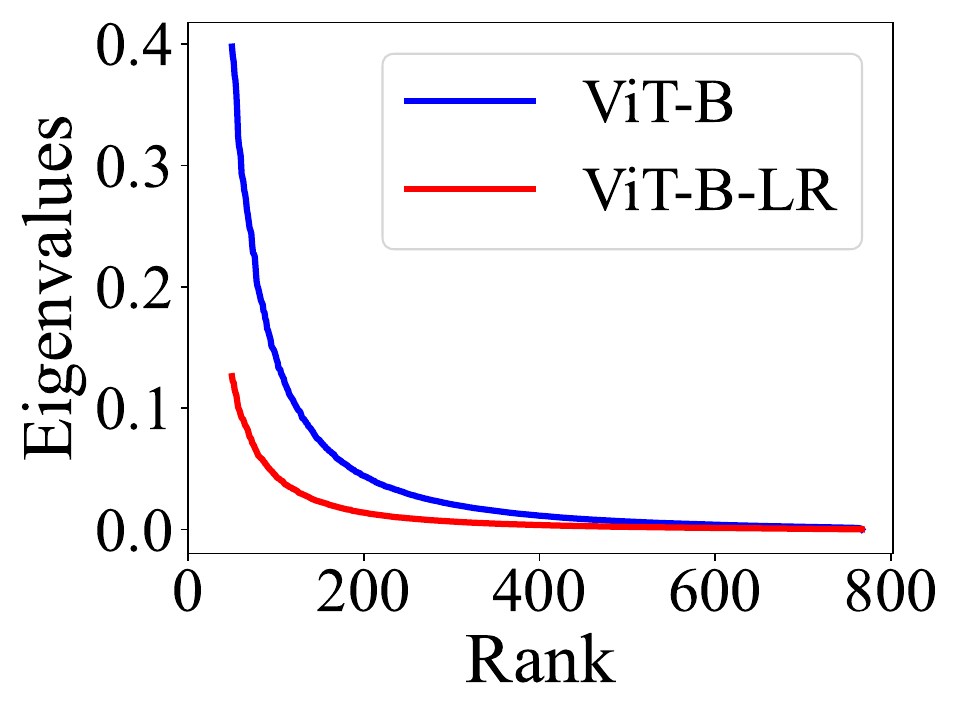}
    }
    \hspace{-0.2cm}
     \subfigure[COVIDx]{
        \includegraphics[width=0.3\textwidth]{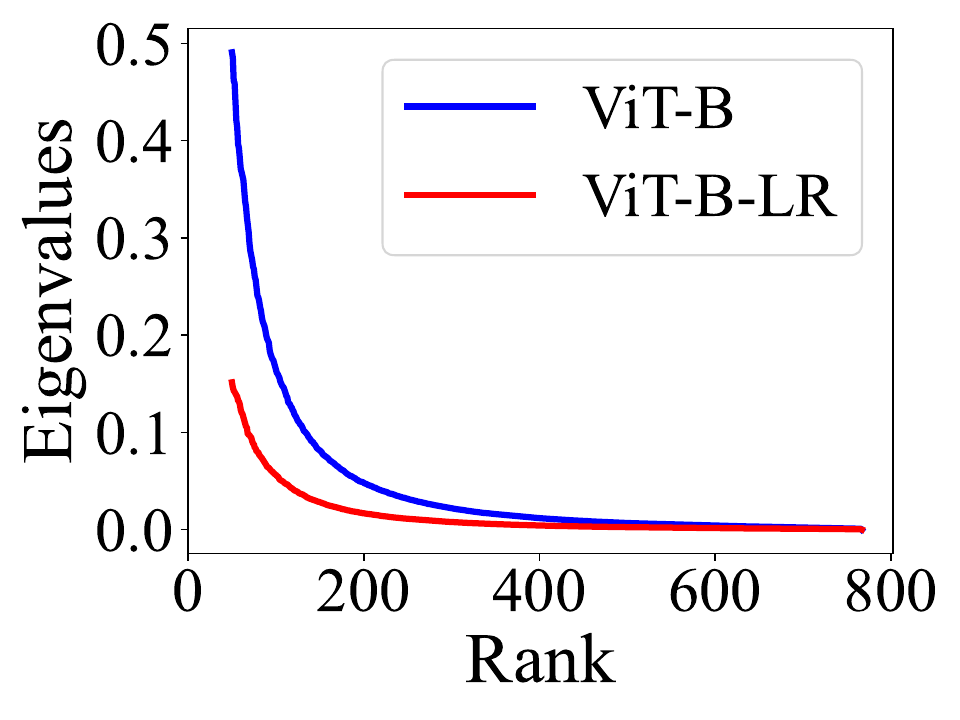}
    }
    \hspace{-0.2cm}
     \subfigure[CheXpert]{
        \includegraphics[width=0.3\textwidth]{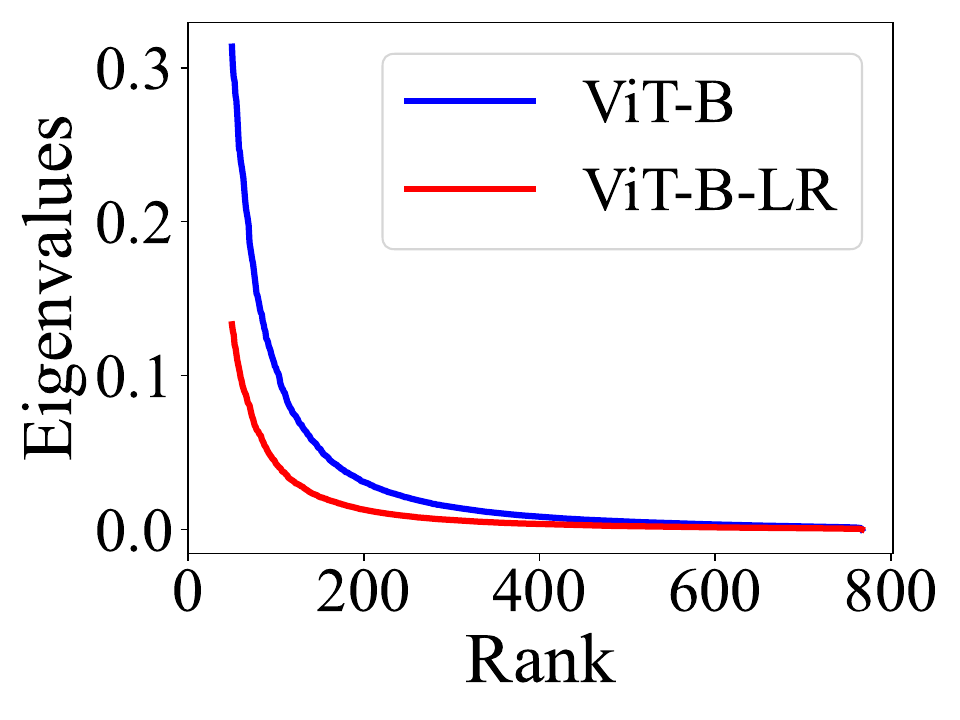}
    }
    \caption{Eigenvalues comparison between ViT-B-LR and ViT-B on ChestX-ray14, COVIDx, and CheXpert.}
     \label{fig:eigenvalues}
\end{figure*}

\subsection{Experiments in Small Data Regimes}
\label{subsection:low_data_regime}

\noindent\textbf{Experimental setup.} We explore the effectiveness of low-rank features learned in scenarios with limited data availability, which is particularly significant given the challenges in acquiring high-quality data annotations in the medical imaging domain. We expect that LRFL models can demonstrate improved performance in such situations due to our theoretical guarantee of the better generalization capability of LRFL. We randomly select  5\%, 10\%, 15\%, 20\%, 25\%, and 50\% of training data from the NIH ChestX-ray14 dataset and then fine-tune the base model using its default training configurations. We then train LRFL models for $20$ epochs.

\smallskip\noindent\textbf{Results and analysis.}
As depicted in Table \ref{table:low-data-regimes}, our LRFL models consistently outperform their corresponding base methods across all data subsets, including 5\%, 10\%, 15\%, 20\%, 25\%, and 50\% on the NIH ChestX-ray14 dataset. Notably, the average improvement in performance is more substantial for the 5\% data subset compared to the remaining subsets. For instance, ViT-B-LR exhibits a remarkable improvement of 1.05\% for the 5\% data subset, which significantly surpasses the improvements of 0.15\%, 0.06\%, 0.06\%, 0.09\% and 0.11\% observed for the 10\%, 15\%, 20\%, 25\% and 50\% training data subsets, respectively. These findings are consistent with our expectations, showcasing the strong generalization capability of LRFL models in mitigating over-fitting issues with limited data. In conclusion, our findings in the low-data regimes demonstrate the superiority of our LRFL in delivering more generalizable and robust representations for tasks with limited data availability, thereby contributing to the reduction of annotation costs. Models of this nature hold substantial value across various medical tasks, particularly because annotating medical images tends to be challenging and necessitates specialized expertise.

\subsection{Ablation study}
\label{subsection:ablation_study}
\subsubsection{Study on the Kernel Eigenvalues and Kernel Complexity}
In this section, we compare the kernel eigenvalues and kernel complexity of ViT-B-LR and ViT-B on ChestX-ray14, COVIDx, and CheXpert. The eigenvalues of ViT-B-LR and ViT-B on ChestX-ray14, COVIDx, and CheXpert are shown in Figure~\ref{fig:eigenvalues}. The kernel complexity of ViT-B-LR and ViT-B on ChestX-ray14, COVIDx, and CheXpert are shown in Table~\ref{table:kernel_complexity}.
\begin{table}[!htbp]
\center
\begin{center}
\resizebox{1\columnwidth}{!}{
\begin{tabular}{|c|cc|cc|cc|}
\hline
\multirow{2}{*}{Method} & \multicolumn{2}{c|}{ChestX-ray14}     & \multicolumn{2}{c|}{COVIDx}           & \multicolumn{2}{c|}{CheXpert}         \\ \cline{2-7}
                        & \multicolumn{1}{c|}{Complexity} & h   & \multicolumn{1}{c|}{Complexity} & h   & \multicolumn{1}{c|}{Complexity} & h   \\ \hline
ViT-B                   & \multicolumn{1}{c|}{0.0101}     & 465 & \multicolumn{1}{c|}{0.0207}     & 303 & \multicolumn{1}{c|}{0.0040}     & 766 \\ \hline
ViT-B-LR                & \multicolumn{1}{c|}{0.0076}     & 262 & \multicolumn{1}{c|}{0.0155}     & 187 & \multicolumn{1}{c|}{0.0038}     & 389 \\ \hline
\end{tabular}
}
\caption{Kernel complexity comparison between ViT-B-LR and ViT-B on ChestX-ray14, COVIDx, and CheXpert.}
\label{table:kernel_complexity}
\end{center}
\end{table}

\subsubsection{Study on the Rank in Low-rank Feature Learning}
\begin{figure}[!htbp]
\begin{center}{
\hspace{-6mm}\includegraphics[width=0.8\columnwidth]{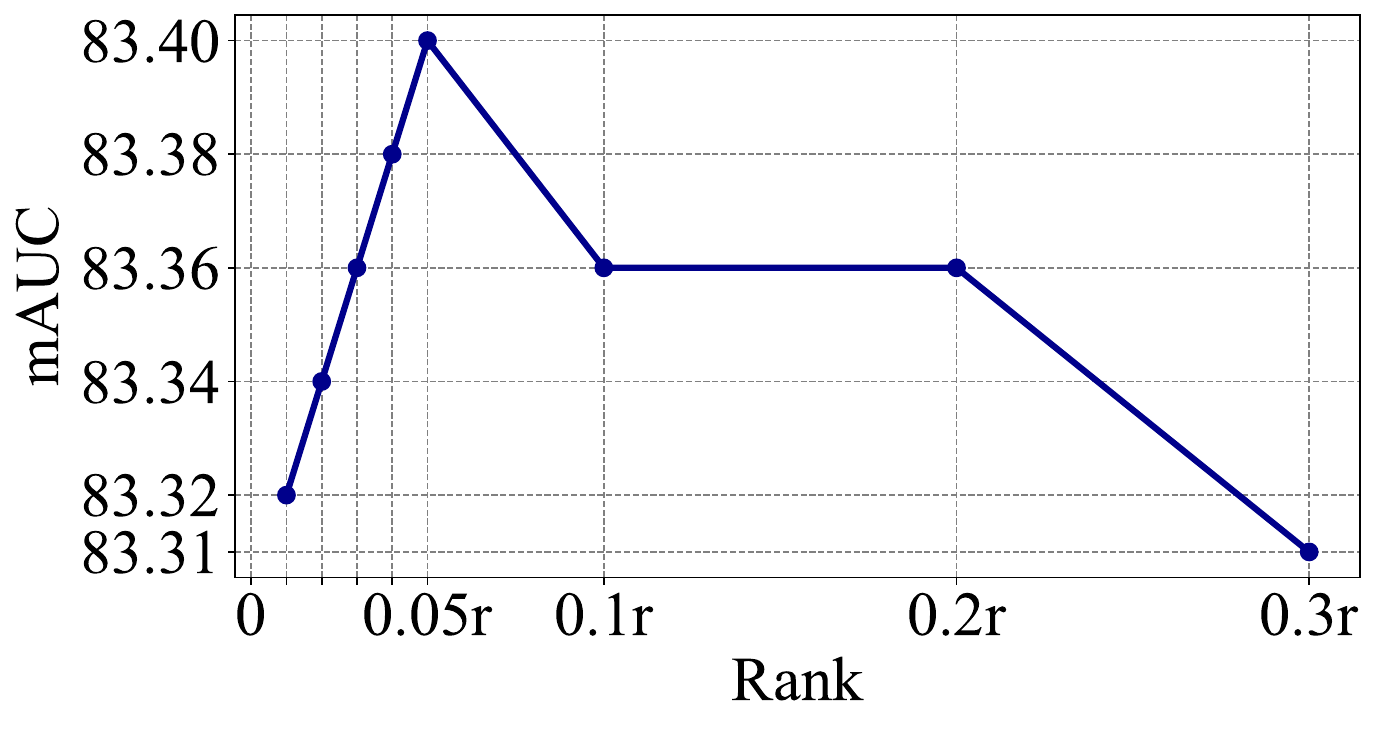}
}
\caption{The relationship between mAUC and rank $T$ of ViT-B-LR on ChestX-ray14.}
\label{fig:aucvrank}
\end{center}
\end{figure}

In this section, we conduct an ablation study to investigate the impact of the rank $T$ on the performance of our LRFL model on the NIH-ChestX-ray14 dataset for the ViT-B/16 model. In Figure \ref{fig:aucvrank}, we present the performance change with respect to different values for the rank $T$, with $T$ varying from $0.01r$ to $0.3r$, where $r = \min\set{n, d}$. It can be observed from this figure that our LRFL model often delivers aAUC higher than $83\%$ for most values for $T$.

\begin{table*}[!htpb]
\center
\begin{center}

\resizebox{1\linewidth}{!}{%

\begin{tabular}{|c|c|c|c|c|c|c|c|c|c|c|c|c|c|}
\hline
\multirow{4}{*}{Pre-training Dataset} & \multirow{4}{*}{Model} & \multicolumn{12}{c|}{Label Fractions}                                       \\
\cline{3-14}
                                       &                        & \multicolumn{2}{c|}{5\% }  & \multicolumn{2}{c|}{10\% }  &\multicolumn{2}{c|}{15\% }  & \multicolumn{2}{c|}{20\% }  &\multicolumn{2}{c|}{25\% }  & \multicolumn{2}{c|}{50\% }   \\
\cline{3-14}
                                       &                        &  Rank     &  mAUC &Rank     &  mAUC &Rank     &  mAUC &Rank     &  mAUC & Rank     &  mAUC  & Rank     &  mAUC\\
\hline
\multirow{2}{*}{X-rays(0.3M)}         & ViT-S            & -& 61.22 &  -& 73.19 & - & 76.99 & - & 78.65 & - & 79.57 & -& 81.20   \\
\cline{2-14}
                                       & ViT-S-LR(Ours)        &0.05$r$ & 61.81 &0.2$r$  & 73.84 & 0.04$r$ & 77.21 & 0.04$r$ & 78.86 & 0.05$r$ & 79.65 & 0.05$r$ & 81.35   \\
\hline
\multirow{2}{*}{X-rays(0.5M)}         & ViT-B            &-& 70.71 & -& 78.67 &- & 79.99 & -& 80.59 &-& 81.13 & -& 82.19   \\
\cline{2-14}
                                       & ViT-B-LR (Ours)        &0.05$r$ & 71.76 & 0.2$r$ & 78.82 & 0.2$r$ & 80.05 & 0.1r & 80.65 & 0.05$r$& 81.22 &0.05$r$ & 82.30   \\
\hline
\end{tabular}

}
\caption{The table evaluates the performance of various models under low data regimes on the NIH ChestX-rays14 dataset. Models trained with low-rank features effectively combat overfitting in scenarios with limited data availability, thereby enhancing the quality of representations for downstream tasks.}
\label{table:low-data-regimes}
\end{center}
\end{table*}

\begin{table*}[!htbp]
\centering
\resizebox{1\textwidth}{!}{%
\begin{tabular}{|c|c|c|c|c|c|c|}
\hline
    \multirow{2}{*}{Model} & \multicolumn{6}{c|}{mAUC} \\ \cline{2-7}
    & Base Model & Fine-tuning & Mix-up~\citep{ZhangCDL18} & Label Smoothing~\citep{muller2019does} &EMA~\citep{rw2019timm}  & LRFL \\ \hline
    ViT-S & 82.27 & 82.26 &82.09& 82.24&82.26& 82.70 \\ \hline
    ViT-B & \underline{83.00} & \underline{83.00} & 82.37 & 82.99 & 82.98 & \textbf{83.40} \\ \hline
\end{tabular}%
}
\caption{Comparison of fine-tuning strategies on NIH ChestX-ray14.}
\label{tab:diff-finetuning}
\end{table*}
\subsubsection{Exploring Fine-tuning Strategies}
Our LRFL method learns low-rank features by leveraging models pre-trained on the target dataset. In this section, we conduct an ablation study to investigate the significance of low-rank regularization in the fine-tuning process. In Table \ref{tab:diff-finetuning}, we present a comparative analysis of low-rank regularization compared with several performance-enhancing techniques, including mix-up~\citep{ZhangCDL18}, label smoothing~\citep{muller2019does}, and EMA~\citep{rw2019timm}. We also perform an experiment by fine-tuning without low-rank regularization and other tricks, which serves as a baseline for studying the effects of fine-tuning strategies. All models undergo equivalent training epochs to ensure a fair comparison. It can be observed that the LRFL models achieve the highest performance improvement compared to all other approaches. Unlike natural images, results in Table~\ref{tab:diff-finetuning} show that applying mix-up, label smoothing, or EMA to the NIH ChestX-ray dataset lead to performance drops. In addition, it is observed that fine-tuning models pre-trained on the target dataset without low-rank regularization does not lead to performance improvements compared to fine-tuning with low-rank regularization. For example, the original ViT-S \citep{xiao2023delving} achieves a Mean AUC of $82.27 \%$ on NIH Chest Xray-14. Fine-tuning this model for $20$ epochs without low-rank regularization leads to a mean AUC of $82.26 \%$. In contrast, fine-tuning with low-rank regularization for 75 epochs leads to a mean AUC of $83.40 \%$. We observe similar results for all models based on low-rank features, which demonstrates the significance of LRFL.



\section{Conclusion}
In this paper, we propose a novel Low-Rank Feature Learning (LRFL) method for thorax disease classification, which can effectively reduce the adverse effect of noise and background, or non-disease areas, on the radiographic images for disease classification. Being universally applicable to the training of all neural networks, LRFL is both empirically motivated by the low frequency property and theoretically motivated by our sharp generalization bound for neural networks with low-rank features. Extensive experimental results on thorax disease datasets, including NIH-ChestX-ray, COVIDx, and CheXpert, demonstrate the superior performance of LRFL in terms of mAUC and classification accuracy.
\bibliography{ref}

\newpage

\onecolumn


\appendix


\section{Proofs}

\begin{proof}[\textup{\bf Proof of Theorem~\ref{theorem:optimization-linear-kernel}}]
It can be verified that at the $t$-th iteration of gradient descent for $t \ge 1$, we have
\bal\label{eq:optimization-linear-kernel-seg1}
\bW^{(t)} = \bW^{(t-1)} - \frac {\eta}{n} \bF^{\top}  \pth{\bF \bW^{(t-1)} - \bY}.
\eal
It follows by (\ref{eq:optimization-linear-kernel-seg1}) that
\bal\label{eq:optimization-linear-kernel-seg2}
\bF \bW^{(t)} &=  \bF \bW^{(t-1)} - \eta \bK_n \pth{\bF \bW^{(t-1)} - \bY} \nonumber \\
&=\bF \bW^{(t-1)} - \eta \bK_n \pth{\bF \bW^{(t-1)} - \bar \bY},
\eal
where $\bK =  \bF \bF^{\top}$, $\bar \bY = \bU^{\bar r} {\bU^{\bar r}}^{\top} \bY$.

We define $\bF(\bW,t) \defeq \bF \bW^{(t)}$, then it follows by (\ref{eq:optimization-linear-kernel-seg2}) that
\bals
\bF(\bW,t) - \bar \bY = \pth{\bI_n - \eta \bK_n } \pth{\bF(\bW,t) - \bar \bY},
\eals
which indicates that
\bals
\bF(\bW,t) - \bar \bY = \pth{\bI_n - \eta \bK_n }^t \pth{\bF(\bW,0) - \bar \bY} = - \pth{\bI_n - \eta \bK_n }^t \bar \bY,
\eals
and
\bal\label{eq:optimization-linear-kernel-full-loss}
\fnorm{\bF(\bW,t) - \bY} \le \fnorm{\bY - \bar \bY} + \pth{1-\eta \hat\lambda_r}^t \fnorm{\bY} \le \pth{1-\eta \hat\lambda_r}^t \fnorm{\bar \bY}
\le \pth{1-\eta \hat\lambda_r}^t \fnorm{\bY}.
\eal
As a result of (\ref{eq:optimization-linear-kernel-full-loss}), by using the proof of ~\citep[Theorem 3.3]{bartlett2005}, for every $x > 0$,
with probability at least $1-\exp(-x)$,
\bal
L_{\cD}(\Net_{\bW}) \le \fnorm{\bY - \bar \bY} + c_1\pth{1-\eta \hat\lambda_r}^{2t} \fnorm{\bY}^2
+ c_2 \min\limits_{h \in [0,r]} \pth{\frac hn + \sqrt{\frac 1n
\sum\limits_{i=h+1}^r \hat \lambda_i}} + \frac{c_3x}{n}.
\eal
\end{proof}

\section{More Experiment Results}
\begin{figure}[!htb]
\begin{center}
\resizebox{0.75\columnwidth}{!}{\includegraphics[width=1\textwidth]{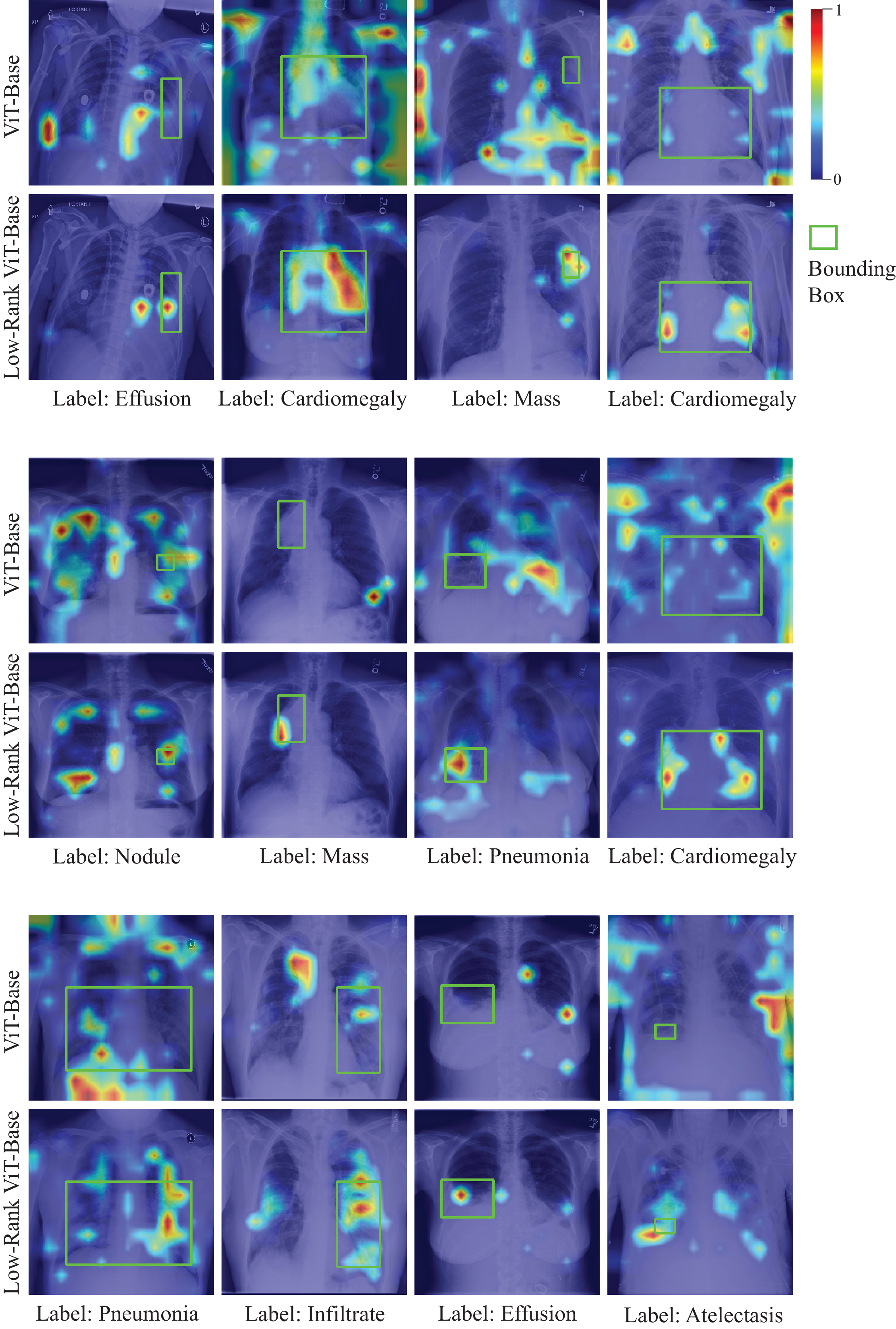}
}
\caption{Grad-CAM visualization results. The figures in the first row are the visualization results of ViT-Base, and the figures in the second row are the visualization results of Low-Rank ViT-Base.}
\label{fig:sup_grad-cam}
\end{center}
\end{figure}
\subsection{Additional Grad-CAM Visualization Results}
In this section, we show more grad-cam visualization results. We visualize the parts in the input images that are responsible for the predictions of the ground-truth disease label for base models and low-rank models. Examples of visualization results in Figure~\ref{fig:sup_grad-cam} show that our low-rank models usually focus more on the areas inside the bounding box associated with the labeled disease. In contrast, the base models also focus on the areas outside the bounding box or even areas in the background.

\end{document}

%% file: symbol_tit.tex
\newcounter{optproblem}

\newtheoremstyle{mytheoremstyle} 
    {\topsep}                    
    {\topsep}                    
    {\normalfont}                
    {}                           
    {\bfseries}                   
    {.}                          
    {.5em}                       
    {} 

\theoremstyle{mytheoremstyle}
\newtheorem{theorem}{Theorem}[section]
\newtheorem{remark}[theorem]{Remark}

\newtheorem*{theorem*}{Theorem}
\newtheorem*{lemma*}{Lemma}
\newtheorem*{remark*}{Remark}

\theoremstyle{remark}

\DeclareMathAlphabet{\pazocal}{OMS}{zplm}{m}{n}
\DeclareMathAlphabet{\mathpzc}{OMS}{pzc}{m}{it}

\setlist[itemize]{leftmargin=*}

\renewcommand{\hat}{\widehat}

\newcommand{\bfm}[1]{\ensuremath{\mathbf{#1}}}
\newcommand{\bfsym}[1]{\ensuremath{\boldsymbol{#1}}}

   \def\bA{\bfm A}

  \def\bF{\bfm F}

   \def\bI{\bfm I}  
     
   \def\bK{\bfm K}

\def\bp{\bfm p}     
     
     \def\RR{\mathbb{R}}

\def\bu{\bfm u}   \def\bU{\bfm U}  
\def\bv{\bfm v}   \def\bV{\bfm V}  
   \def\bW{\bfm W}  
\def\bx{\bfm x}     
\def\by{\bfm y}   \def\bY{\bfm Y}

 \def\cB{{\cal  B}}
 
 \def\cD{{\cal  D}}

 \def\cL{{\cal  L}}

 \def\cW{{\cal  W}}

\def\bSigma{\bfsym \Sigma}

\def\+#1{\mathcal{#1}}
\def\-#1{\textup{#1}}

\def\set#1{\left\{ #1 \right\}}
\def\pth#1{\left( #1 \right)}
\def\bth#1{\left[ #1 \right]}
\def\abth#1{\left | #1 \right |}

\def\defeq {\coloneqq}

\newcommand{\La}{\left\langle\kern-0.64ex\left\langle}
\newcommand{\Ra}{\right\rangle\kern-0.64ex\right\rangle}

\def\Norm#1#2{{\left\vert\kern-0.4ex\left\vert\kern-0.4ex\left\vert #1
    \right\vert\kern-0.4ex\right\vert\kern-0.4ex\right\vert}_{#2}}
\def\norm#1#2{{\left\|#1\right\|}_{#2}}

\def\ltwonorm#1{\norm{#1}{2}}
\def\fnorm#1{\norm{#1}{\textup{F}}}
\def\supnorm#1{\norm{#1}{\infty}}

\newcommand{\1}{{\rm 1}\kern-0.25em{\rm I}}
\def\indict#1{{\rm 1}\kern-0.25em{\rm I}_{\set{#1}}}

\def\set#1{\left\{#1\right\}}

\def \E {\mathbb{E}}
\def\Expect#1#2{\E_{#1}\left[#2\right]}

\newcommand{\beq}{\begin{equation}}
\newcommand{\eeq}{\end{equation}}
\newcommand{\beqa}{\begin{eqnarray}}
\newcommand{\eeqa}{\end{eqnarray}}
\newcommand{\beqas}{\begin{eqnarray*}}
\newcommand{\eeqas}{\end{eqnarray*}}
\def\bal#1\eal{\begin{align}#1\end{align}}
\def\bals#1\eals{\begin{align*}#1\end{align*}}
\def\bsal#1\esal{\begin{small}\begin{align}#1\end{align}\end{small}}
\def\bsals#1\esals{\begin{small}\begin{align*}#1\end{align*}\end{small}}
\def\bsfal#1\esfal{\begin{small}\begin{flalign}#1\end{flalign}\end{small}}

%% file: low_rank_medical_imaging.bbl
\begin{thebibliography}{109}
\providecommand{\natexlab}[1]{#1}
\providecommand{\url}[1]{\texttt{#1}}
\expandafter\ifx\csname urlstyle\endcsname\relax
  \providecommand{\doi}[1]{doi: #1}\else
  \providecommand{\doi}{doi: \begingroup \urlstyle{rm}\Url}\fi

\bibitem[Alex et~al.(2017)Alex, Vaidhya, Thirunavukkarasu, Kesavadas, and
  Krishnamurthi]{alex2017semisupervised}
Varghese Alex, Kiran Vaidhya, Subramaniam Thirunavukkarasu, Chandrasekharan
  Kesavadas, and Ganapathy Krishnamurthi.
\newblock Semisupervised learning using denoising autoencoders for brain lesion
  detection and segmentation.
\newblock \emph{Journal of Medical Imaging}, 4\penalty0 (4):\penalty0 041311,
  2017.

\bibitem[Allaouzi and Ahmed(2019)]{allaouzi2019novel}
Imane Allaouzi and Mohamed~Ben Ahmed.
\newblock A novel approach for multi-label chest x-ray classification of common
  thorax diseases.
\newblock \emph{IEEE Access}, 7:\penalty0 64279--64288, 2019.

\bibitem[Azizi et~al.(2022)Azizi, Culp, Freyberg, Mustafa, Baur, Kornblith,
  Chen, MacWilliams, Mahdavi, Wulczyn, et~al.]{azizi2022robust}
Shekoofeh Azizi, Laura Culp, Jan Freyberg, Basil Mustafa, Sebastien Baur, Simon
  Kornblith, Ting Chen, Patricia MacWilliams, S~Sara Mahdavi, Ellery Wulczyn,
  et~al.
\newblock Robust and efficient medical imaging with self-supervision.
\newblock \emph{arXiv preprint arXiv:2205.09723}, 2022.

\bibitem[Bai et~al.(2021)Bai, Mei, Yuille, and Xie]{bai2021transformers}
Yutong Bai, Jieru Mei, Alan~L Yuille, and Cihang Xie.
\newblock Are transformers more robust than cnns?
\newblock \emph{Advances in Neural Information Processing Systems},
  34:\penalty0 26831--26843, 2021.

\bibitem[Baltruschat et~al.(2019)Baltruschat, Nickisch, Grass, Knopp, and
  Saalbach]{baltruschat2019comparison}
Ivo~M Baltruschat, Hannes Nickisch, Michael Grass, Tobias Knopp, and Axel
  Saalbach.
\newblock Comparison of deep learning approaches for multi-label chest x-ray
  classification.
\newblock \emph{Scientific reports}, 9\penalty0 (1):\penalty0 1--10, 2019.

\bibitem[Bao et~al.(2021)Bao, Dong, and Wei]{bao2021beit}
Hangbo Bao, Li~Dong, and Furu Wei.
\newblock Beit: Bert pre-training of image transformers.
\newblock \emph{arXiv preprint arXiv:2106.08254}, 2021.

\bibitem[Bartlett et~al.(2005)Bartlett, Bousquet, and Mendelson]{bartlett2005}
Peter~L. Bartlett, Olivier Bousquet, and Shahar Mendelson.
\newblock Local rademacher complexities.
\newblock \emph{Ann. Statist.}, 33\penalty0 (4):\penalty0 1497--1537, 08 2005.

\bibitem[Cai et~al.(2023)Cai, Gan, and Han]{cai2022efficientvit}
Han Cai, Chuang Gan, and Song Han.
\newblock Efficientvit: Enhanced linear attention for high-resolution
  low-computation visual recognition.
\newblock In \emph{Proceedings of the IEEE/CVF International Conference on
  Computer Vision}, 2023.

\bibitem[{\c{C}}all{\i} et~al.(2021){\c{C}}all{\i}, Sogancioglu, van Ginneken,
  van Leeuwen, and Murphy]{ccalli2021deep}
Erdi {\c{C}}all{\i}, Ecem Sogancioglu, Bram van Ginneken, Kicky~G van Leeuwen,
  and Keelin Murphy.
\newblock Deep learning for chest x-ray analysis: A survey.
\newblock \emph{Medical Image Analysis}, 72:\penalty0 102125, 2021.

\bibitem[Caron et~al.(2020)Caron, Misra, Mairal, Goyal, Bojanowski, and
  Joulin]{caron2020unsupervised}
Mathilde Caron, Ishan Misra, Julien Mairal, Priya Goyal, Piotr Bojanowski, and
  Armand Joulin.
\newblock Unsupervised learning of visual features by contrasting cluster
  assignments.
\newblock \emph{arXiv preprint arXiv:2006.09882}, 2020.

\bibitem[Chandra and Verma(2020)]{chandra2020analysis}
Tej~Bahadur Chandra and Kesari Verma.
\newblock Analysis of quantum noise-reducing filters on chest x-ray images: A
  review.
\newblock \emph{Measurement}, 153:\penalty0 107426, 2020.

\bibitem[Chefer et~al.(2021)Chefer, Gur, and Wolf]{chefer2021transformer}
Hila Chefer, Shir Gur, and Lior Wolf.
\newblock Transformer interpretability beyond attention visualization.
\newblock In \emph{Proceedings of the IEEE/CVF Conference on Computer Vision
  and Pattern Recognition}, pages 782--791, 2021.

\bibitem[Chen et~al.(2021{\natexlab{a}})Chen, Wang, Guo, Xu, Deng, Liu, Ma, Xu,
  Xu, and Gao]{chen2021pre}
Hanting Chen, Yunhe Wang, Tianyu Guo, Chang Xu, Yiping Deng, Zhenhua Liu, Siwei
  Ma, Chunjing Xu, Chao Xu, and Wen Gao.
\newblock Pre-trained image processing transformer.
\newblock In \emph{Proceedings of the IEEE/CVF conference on computer vision
  and pattern recognition}, pages 12299--12310, 2021{\natexlab{a}}.

\bibitem[Chen et~al.(2021{\natexlab{b}})Chen, Lu, Yu, Luo, Adeli, Wang, Lu,
  Yuille, and Zhou]{chen2021transunet}
Jieneng Chen, Yongyi Lu, Qihang Yu, Xiangde Luo, Ehsan Adeli, Yan Wang, Le~Lu,
  Alan~L Yuille, and Yuyin Zhou.
\newblock Transunet: Transformers make strong encoders for medical image
  segmentation.
\newblock \emph{arXiv preprint arXiv:2102.04306}, 2021{\natexlab{b}}.

\bibitem[Chen et~al.(2019)Chen, Bentley, Mori, Misawa, Fujiwara, and
  Rueckert]{chen2019self}
Liang Chen, Paul Bentley, Kensaku Mori, Kazunari Misawa, Michitaka Fujiwara,
  and Daniel Rueckert.
\newblock Self-supervised learning for medical image analysis using image
  context restoration.
\newblock \emph{Medical image analysis}, 58:\penalty0 101539, 2019.

\bibitem[Chen et~al.(2020{\natexlab{a}})Chen, Radford, Child, Wu, and
  Jun]{chen2020imagegpt}
Mark Chen, Alec Radford, Rewon Child, Jeff Wu, and Heewoo Jun.
\newblock Generative pretraining from pixels.
\newblock \emph{Advances in Neural Information Processing Systems},
  2020{\natexlab{a}}.

\bibitem[Chen et~al.(2020{\natexlab{b}})Chen, Kornblith, Norouzi, and
  Hinton]{chen2020simple}
Ting Chen, Simon Kornblith, Mohammad Norouzi, and Geoffrey Hinton.
\newblock A simple framework for contrastive learning of visual
  representations.
\newblock \emph{arXiv preprint arXiv:2002.05709}, 2020{\natexlab{b}}.

\bibitem[Chen et~al.(2020{\natexlab{c}})Chen, Fan, Girshick, and
  He]{chen2020improved}
Xinlei Chen, Haoqi Fan, Ross Girshick, and Kaiming He.
\newblock Improved baselines with momentum contrastive learning.
\newblock \emph{arXiv preprint arXiv:2003.04297}, 2020{\natexlab{c}}.

\bibitem[Cui et~al.(2019)Cui, Liu, and Huang]{cui2019pulmonary}
Hejie Cui, Xinglong Liu, and Ning Huang.
\newblock Pulmonary vessel segmentation based on orthogonal fused u-net++ of
  chest ct images.
\newblock In \emph{International Conference on Medical Image Computing and
  Computer-Assisted Intervention}, pages 293--300. Springer, 2019.

\bibitem[Deng et~al.(2009)Deng, Dong, Socher, Li, Li, and
  Fei-Fei]{deng2009imagenet}
Jia Deng, Wei Dong, Richard Socher, Li-Jia Li, Kai Li, and Li~Fei-Fei.
\newblock Imagenet: A large-scale hierarchical image database.
\newblock In \emph{Proceedings of the IEEE Conference on Computer Vision and
  Pattern Recognition}, pages 248--255. IEEE, 2009.

\bibitem[Ding et~al.(2018)Ding, Long, Zhang, and Fessler]{ding2018statistical}
Qiaoqiao Ding, Yong Long, Xiaoqun Zhang, and Jeffrey~A Fessler.
\newblock Statistical image reconstruction using mixed poisson-gaussian noise
  model for x-ray ct.
\newblock \emph{arXiv preprint arXiv:1801.09533}, 2018.

\bibitem[Ding et~al.(2022)Ding, Zhang, Han, and Ding]{ding2022scaling}
Xiaohan Ding, Xiangyu Zhang, Jungong Han, and Guiguang Ding.
\newblock Scaling up your kernels to 31x31: Revisiting large kernel design in
  cnns.
\newblock In \emph{Proceedings of the IEEE/CVF Conference on Computer Vision
  and Pattern Recognition}, pages 11963--11975, 2022.

\bibitem[Dosovitskiy et~al.(2020)Dosovitskiy, Beyer, Kolesnikov, Weissenborn,
  Zhai, Unterthiner, Dehghani, Minderer, Heigold, Gelly,
  et~al.]{dosovitskiy2020image}
Alexey Dosovitskiy, Lucas Beyer, Alexander Kolesnikov, Dirk Weissenborn,
  Xiaohua Zhai, Thomas Unterthiner, Mostafa Dehghani, Matthias Minderer, Georg
  Heigold, Sylvain Gelly, et~al.
\newblock An image is worth 16x16 words: Transformers for image recognition at
  scale.
\newblock In \emph{ICLR}, 2020.

\bibitem[Falk et~al.(2018)Falk, Mai, Bensch, {\c{C}}i{\c{c}}ek, Abdulkadir,
  Marrakchi, B{\"o}hm, Deubner, J{\"a}ckel, Seiwald, et~al.]{falk2018u}
Thorsten Falk, Dominic Mai, Robert Bensch, {\"O}zg{\"u}n {\c{C}}i{\c{c}}ek,
  Ahmed Abdulkadir, Yassine Marrakchi, Anton B{\"o}hm, Jan Deubner, Zoe
  J{\"a}ckel, Katharina Seiwald, et~al.
\newblock U-net: deep learning for cell counting, detection, and morphometry.
\newblock \emph{Nature methods}, page~1, 2018.

\bibitem[Falk et~al.(2019)Falk, Mai, Bensch, {\c{C}}i{\c{c}}ek, Abdulkadir,
  Marrakchi, B{\"o}hm, Deubner, J{\"a}ckel, Seiwald, et~al.]{falk2019u}
Thorsten Falk, Dominic Mai, Robert Bensch, {\"O}zg{\"u}n {\c{C}}i{\c{c}}ek,
  Ahmed Abdulkadir, Yassine Marrakchi, Anton B{\"o}hm, Jan Deubner, Zoe
  J{\"a}ckel, Katharina Seiwald, et~al.
\newblock U-net: deep learning for cell counting, detection, and morphometry.
\newblock \emph{Nature methods}, 16\penalty0 (1):\penalty0 67--70, 2019.

\bibitem[Feng et~al.(2020)Feng, Zhou, Gotway, and Liang]{feng2020parts2whole}
Ruibin Feng, Zongwei Zhou, Michael~B Gotway, and Jianming Liang.
\newblock Parts2whole: Self-supervised contrastive learning via reconstruction.
\newblock In \emph{Domain Adaptation and Representation Transfer, and
  Distributed and Collaborative Learning}, pages 85--95. Springer, 2020.

\bibitem[Gao et~al.(2019)Gao, Huo, Bao, Tang, Antic, Epstein, Balar, Deppen,
  Paulson, Sandler, et~al.]{gao2019distanced}
Riqiang Gao, Yuankai Huo, Shunxing Bao, Yucheng Tang, Sanja~L Antic, Emily~S
  Epstein, Aneri~B Balar, Steve Deppen, Alexis~B Paulson, Kim~L Sandler, et~al.
\newblock Distanced lstm: time-distanced gates in long short-term memory models
  for lung cancer detection.
\newblock In \emph{Machine Learning in Medical Imaging: 10th International
  Workshop, MLMI 2019, Held in Conjunction with MICCAI 2019, Shenzhen, China,
  October 13, 2019, Proceedings 10}, pages 310--318. Springer, 2019.

\bibitem[Grill et~al.(2020)Grill, Strub, Altch{\'e}, Tallec, Richemond,
  Buchatskaya, Doersch, Pires, Guo, Azar, et~al.]{grill2020bootstrap}
Jean-Bastien Grill, Florian Strub, Florent Altch{\'e}, Corentin Tallec,
  Pierre~H Richemond, Elena Buchatskaya, Carl Doersch, Bernardo~Avila Pires,
  Zhaohan~Daniel Guo, Mohammad~Gheshlaghi Azar, et~al.
\newblock Bootstrap your own latent: A new approach to self-supervised
  learning.
\newblock \emph{arXiv preprint arXiv:2006.07733}, 2020.

\bibitem[Guan and Huang(2018)]{guan2018multi}
Qingji Guan and Yaping Huang.
\newblock Multi-label chest x-ray image classification via category-wise
  residual attention learning.
\newblock \emph{Pattern Recognition Letters}, 2018.

\bibitem[Guendel et~al.(2018)Guendel, Grbic, Georgescu, Liu, Maier, and
  Comaniciu]{guendel2018learning}
Sebastian Guendel, Sasa Grbic, Bogdan Georgescu, Siqi Liu, Andreas Maier, and
  Dorin Comaniciu.
\newblock Learning to recognize abnormalities in chest x-rays with
  location-aware dense networks.
\newblock In \emph{Iberoamerican Congress on Pattern Recognition}, pages
  757--765. Springer, 2018.

\bibitem[Haghighi et~al.(2022)Haghighi, Taher, Gotway, and
  Liang]{haghighi2022dira}
Fatemeh Haghighi, Mohammad Reza~Hosseinzadeh Taher, Michael~B Gotway, and
  Jianming Liang.
\newblock Dira: Discriminative, restorative, and adversarial learning for
  self-supervised medical image analysis.
\newblock In \emph{Proceedings of the IEEE/CVF Conference on Computer Vision
  and Pattern Recognition}, pages 20824--20834, 2022.

\bibitem[Hatamizadeh et~al.(2022)Hatamizadeh, Tang, Nath, Yang, Myronenko,
  Landman, Roth, and Xu]{hatamizadeh2022unetr}
Ali Hatamizadeh, Yucheng Tang, Vishwesh Nath, Dong Yang, Andriy Myronenko,
  Bennett Landman, Holger~R Roth, and Daguang Xu.
\newblock Unetr: Transformers for 3d medical image segmentation.
\newblock In \emph{Proceedings of the IEEE/CVF Winter Conference on
  Applications of Computer Vision}, pages 574--584, 2022.

\bibitem[He et~al.(2016)He, Zhang, Ren, and Sun]{he2016deep}
Kaiming He, Xiangyu Zhang, Shaoqing Ren, and Jian Sun.
\newblock Deep residual learning for image recognition.
\newblock In \emph{Proceedings of the IEEE Conference on Computer Vision and
  Pattern Recognition}, pages 770--778, 2016.

\bibitem[He et~al.(2020)He, Fan, Wu, Xie, and Girshick]{he2020momentum}
Kaiming He, Haoqi Fan, Yuxin Wu, Saining Xie, and Ross Girshick.
\newblock Momentum contrast for unsupervised visual representation learning.
\newblock In \emph{Proceedings of the IEEE/CVF Conference on Computer Vision
  and Pattern Recognition}, pages 9729--9738, 2020.

\bibitem[He et~al.(2022)He, Chen, Xie, Li, Doll{\'a}r, and
  Girshick]{he2022masked}
Kaiming He, Xinlei Chen, Saining Xie, Yanghao Li, Piotr Doll{\'a}r, and Ross
  Girshick.
\newblock Masked autoencoders are scalable vision learners.
\newblock In \emph{Proceedings of the IEEE/CVF Conference on Computer Vision
  and Pattern Recognition}, pages 16000--16009, 2022.

\bibitem[Hermoza et~al.(2020)Hermoza, Maicas, Nascimento, and
  Carneiro]{hermoza2020region}
Renato Hermoza, Gabriel Maicas, Jacinto~C Nascimento, and Gustavo Carneiro.
\newblock Region proposals for saliency map refinement for weakly-supervised
  disease localisation and classification.
\newblock In \emph{International Conference on Medical Image Computing and
  Computer-Assisted Intervention}, pages 539--549. Springer, 2020.

\bibitem[Hosseinzadeh~Taher et~al.(2021)Hosseinzadeh~Taher, Haghighi, Feng,
  Gotway, and Liang]{hosseinzadeh2021systematic}
Mohammad~Reza Hosseinzadeh~Taher, Fatemeh Haghighi, Ruibin Feng, Michael~B
  Gotway, and Jianming Liang.
\newblock A systematic benchmarking analysis of transfer learning for medical
  image analysis.
\newblock In \emph{Domain Adaptation and Representation Transfer, and
  Affordable Healthcare and AI for Resource Diverse Global Health}, pages
  3--13. Springer, 2021.

\bibitem[Hu et~al.(2023)Hu, Zhang, Matkovic, Liu, and
  Yang]{hu2023reinforcement}
Mingzhe Hu, Jiahan Zhang, Luke Matkovic, Tian Liu, and Xiaofeng Yang.
\newblock Reinforcement learning in medical image analysis: Concepts,
  applications, challenges, and future directions.
\newblock \emph{Journal of Applied Clinical Medical Physics}, 24\penalty0
  (2):\penalty0 e13898, 2023.

\bibitem[Irvin et~al.(2019)Irvin, Rajpurkar, Ko, Yu, Ciurea-Ilcus, Chute,
  Marklund, Haghgoo, Ball, Shpanskaya, et~al.]{irvin2019chexpert}
Jeremy Irvin, Pranav Rajpurkar, Michael Ko, Yifan Yu, Silviana Ciurea-Ilcus,
  Chris Chute, Henrik Marklund, Behzad Haghgoo, Robyn Ball, Katie Shpanskaya,
  et~al.
\newblock Chexpert: A large chest radiograph dataset with uncertainty labels
  and expert comparison.
\newblock In \emph{Proceedings of the AAAI Conference on Artificial
  Intelligence}, volume~33, pages 590--597, 2019.

\bibitem[Kang et~al.(2021)Kang, Lu, Yuille, and Zhou]{kang2021data}
Mintong Kang, Yongyi Lu, Alan~L Yuille, and Zongwei Zhou.
\newblock Label-assemble: Leveraging multiple datasets with partial labels.
\newblock \emph{In Submission: Thirty-Sixth Conference on Neural Information
  Processing Systems}, 2021.
\newblock URL \url{https://arxiv.org/pdf/2109.12265.pdf}.

\bibitem[Kim et~al.(2021)Kim, Kim, Seo, and Yoon]{Kim_2021_CVPR}
Eunji Kim, Siwon Kim, Minji Seo, and Sungroh Yoon.
\newblock Xprotonet: Diagnosis in chest radiography with global and local
  explanations.
\newblock In \emph{Proceedings of the IEEE/CVF Conference on Computer Vision
  and Pattern Recognition (CVPR)}, pages 15719--15728, June 2021.

\bibitem[Krizhevsky et~al.(2012)Krizhevsky, Sutskever, and
  Hinton]{krizhevsky2012imagenet}
Alex Krizhevsky, Ilya Sutskever, and Geoffrey~E Hinton.
\newblock Imagenet classification with deep convolutional neural networks.
\newblock In \emph{Advances in neural information processing systems}, pages
  1097--1105, 2012.

\bibitem[Lee et~al.(2018)Lee, Lee, and Kang]{lee2018poisson}
Sangyoon Lee, Min~Seok Lee, and Moon~Gi Kang.
\newblock Poisson--gaussian noise analysis and estimation for low-dose x-ray
  images in the nsct domain.
\newblock \emph{Sensors}, 18\penalty0 (4):\penalty0 1019, 2018.

\bibitem[Li et~al.(2022)Li, Chen, Tang, Landman, and Zhou]{li2022transforming}
Jun Li, Junyu Chen, Yucheng Tang, Bennett~A Landman, and S~Kevin Zhou.
\newblock Transforming medical imaging with transformers? a comparative review
  of key properties, current progresses, and future perspectives.
\newblock \emph{arXiv preprint arXiv:2206.01136}, 2022.

\bibitem[Li et~al.(2023)Li, Chen, Tang, Wang, Landman, and
  Zhou]{li2023transforming}
Jun Li, Junyu Chen, Yucheng Tang, Ce~Wang, Bennett~A Landman, and S~Kevin Zhou.
\newblock Transforming medical imaging with transformers? a comparative review
  of key properties, current progresses, and future perspectives.
\newblock \emph{Medical image analysis}, page 102762, 2023.

\bibitem[Li et~al.(2018)Li, Wang, Han, Xue, Wei, Li, and
  Fei-Fei]{li2018thoracic}
Zhe Li, Chong Wang, Mei Han, Yuan Xue, Wei Wei, Li-Jia Li, and Li~Fei-Fei.
\newblock Thoracic disease identification and localization with limited
  supervision.
\newblock In \emph{Proceedings of the IEEE Conference on Computer Vision and
  Pattern Recognition}, pages 8290--8299, 2018.

\bibitem[Lin et~al.(2017{\natexlab{a}})Lin, Doll{\'a}r, Girshick, He,
  Hariharan, and Belongie]{lin2017feature}
Tsung-Yi Lin, Piotr Doll{\'a}r, Ross Girshick, Kaiming He, Bharath Hariharan,
  and Serge Belongie.
\newblock Feature pyramid networks for object detection.
\newblock In \emph{Proceedings of the IEEE Conference on Computer Vision and
  Pattern Recognition}, 2017{\natexlab{a}}.

\bibitem[Lin et~al.(2017{\natexlab{b}})Lin, Goyal, Girshick, He, and
  Doll{\'a}r]{Lin2017a}
Tsung-Yi Lin, Priya Goyal, Ross Girshick, Kaiming He, and Piotr Doll{\'a}r.
\newblock Focal loss for dense object detection.
\newblock In \emph{ICCV}, 2017{\natexlab{b}}.

\bibitem[Liu et~al.(2022{\natexlab{a}})Liu, Tian, Chen, Liu, Belagiannis, and
  Carneiro]{liu2022acpl}
Fengbei Liu, Yu~Tian, Yuanhong Chen, Yuyuan Liu, Vasileios Belagiannis, and
  Gustavo Carneiro.
\newblock Acpl: Anti-curriculum pseudo-labelling for semi-supervised medical
  image classification.
\newblock In \emph{Proceedings of the IEEE/CVF Conference on Computer Vision
  and Pattern Recognition}, pages 20697--20706, 2022{\natexlab{a}}.

\bibitem[Liu et~al.(2021)Liu, Lin, Cao, Hu, Wei, Zhang, Lin, and
  Guo]{liu2021swin}
Ze~Liu, Yutong Lin, Yue Cao, Han Hu, Yixuan Wei, Zheng Zhang, Stephen Lin, and
  Baining Guo.
\newblock Swin transformer: Hierarchical vision transformer using shifted
  windows.
\newblock In \emph{Proceedings of the IEEE/CVF International Conference on
  Computer Vision}, pages 10012--10022, 2021.

\bibitem[Liu et~al.(2022{\natexlab{b}})Liu, Mao, Wu, Feichtenhofer, Darrell,
  and Xie]{liu2022convnet}
Zhuang Liu, Hanzi Mao, Chao-Yuan Wu, Christoph Feichtenhofer, Trevor Darrell,
  and Saining Xie.
\newblock A convnet for the 2020s.
\newblock In \emph{Proceedings of the IEEE/CVF Conference on Computer Vision
  and Pattern Recognition}, pages 11976--11986, 2022{\natexlab{b}}.

\bibitem[Ma et~al.(2020)Ma, Wang, and Hoi]{ma2020multilabel}
Congbo Ma, Hu~Wang, and Steven C.~H. Hoi.
\newblock Multi-label thoracic disease image classification with
  cross-attention networks, 2020.

\bibitem[Ma et~al.(2022)Ma, Hosseinzadeh~Taher, Pang, Islam, Haghighi, Gotway,
  and Liang]{ma2022benchmarking}
DongAo Ma, Mohammad~Reza Hosseinzadeh~Taher, Jiaxuan Pang, Nahid~UI Islam,
  Fatemeh Haghighi, Michael~B Gotway, and Jianming Liang.
\newblock Benchmarking and boosting transformers for medical image
  classification.
\newblock In \emph{MICCAI Workshop on Domain Adaptation and Representation
  Transfer}, pages 12--22. Springer, 2022.

\bibitem[Ma et~al.(2019)Ma, Zhou, Chen, Lu, and Zhao]{ma2019multi}
Yanbo Ma, Qiuhao Zhou, Xuesong Chen, Haihua Lu, and Yong Zhao.
\newblock Multi-attention network for thoracic disease classification and
  localization.
\newblock In \emph{ICASSP 2019-2019 IEEE International Conference on Acoustics,
  Speech and Signal Processing (ICASSP)}, pages 1378--1382. IEEE, 2019.

\bibitem[Manson et~al.(2019)Manson, Ampoh, Fiagbedzi, Amuasi, Flether, and
  Schandorf]{manson2019image}
EN~Manson, V~Atuwo Ampoh, E~Fiagbedzi, JH~Amuasi, JJ~Flether, and C~Schandorf.
\newblock Image noise in radiography and tomography: Causes, effects and
  reduction techniques.
\newblock \emph{Curr. Trends Clin. Med. Imaging}, 2\penalty0 (5):\penalty0
  555620, 2019.

\bibitem[Mao et~al.(2022)Mao, Qi, Chen, Li, Duan, Ye, He, and
  Xue]{mao2022towards}
Xiaofeng Mao, Gege Qi, Yuefeng Chen, Xiaodan Li, Ranjie Duan, Shaokai Ye, Yuan
  He, and Hui Xue.
\newblock Towards robust vision transformer.
\newblock In \emph{Proceedings of the IEEE/CVF Conference on Computer Vision
  and Pattern Recognition}, pages 12042--12051, 2022.

\bibitem[M{\"u}ller et~al.(2019)M{\"u}ller, Kornblith, and
  Hinton]{muller2019does}
Rafael M{\"u}ller, Simon Kornblith, and Geoffrey~E Hinton.
\newblock When does label smoothing help?
\newblock \emph{Advances in neural information processing systems}, 32, 2019.

\bibitem[Paul and Chen(2022)]{paul2022vision}
Sayak Paul and Pin-Yu Chen.
\newblock Vision transformers are robust learners.
\newblock In \emph{Proceedings of the AAAI Conference on Artificial
  Intelligence}, volume~36, pages 2071--2081, 2022.

\bibitem[Pavlova et~al.(2022)Pavlova, Tuinstra, Aboutalebi, Zhao, Gunraj, and
  Wong]{pavlova2022covidx}
Maya Pavlova, Tia Tuinstra, Hossein Aboutalebi, Andy Zhao, Hayden Gunraj, and
  Alexander Wong.
\newblock Covidx cxr-3: a large-scale, open-source benchmark dataset of chest
  x-ray images for computer-aided covid-19 diagnostics.
\newblock \emph{arXiv preprint arXiv:2206.03671}, 2022.

\bibitem[Pham et~al.(2021)Pham, Le, Tran, Ngo, and
  Nguyen]{pham2021interpreting}
Hieu~H Pham, Tung~T Le, Dat~Q Tran, Dat~T Ngo, and Ha~Q Nguyen.
\newblock Interpreting chest x-rays via cnns that exploit hierarchical disease
  dependencies and uncertainty labels.
\newblock \emph{Neurocomputing}, 437:\penalty0 186--194, 2021.

\bibitem[Ronneberger et~al.(2015)Ronneberger, Fischer, and
  Brox]{ronneberger2015u}
Olaf Ronneberger, Philipp Fischer, and Thomas Brox.
\newblock U-net: Convolutional networks for biomedical image segmentation.
\newblock In \emph{International Conference on Medical Image Computing and
  Computer-Assisted Intervention}, pages 234--241. Springer, 2015.

\bibitem[Russakovsky et~al.(2015)Russakovsky, Deng, Su, Krause, Satheesh, Ma,
  Huang, Karpathy, Khosla, Bernstein, et~al.]{russakovsky2015imagenet}
Olga Russakovsky, Jia Deng, Hao Su, Jonathan Krause, Sanjeev Satheesh, Sean Ma,
  Zhiheng Huang, Andrej Karpathy, Aditya Khosla, Michael Bernstein, et~al.
\newblock Imagenet large scale visual recognition challenge.
\newblock \emph{International journal of computer vision}, 115\penalty0
  (3):\penalty0 211--252, 2015.

\bibitem[Selvaraju et~al.(2017)Selvaraju, Cogswell, Das, Vedantam, Parikh, and
  Batra]{selvaraju2017grad}
Ramprasaath~R Selvaraju, Michael Cogswell, Abhishek Das, Ramakrishna Vedantam,
  Devi Parikh, and Dhruv Batra.
\newblock Grad-cam: Visual explanations from deep networks via gradient-based
  localization.
\newblock In \emph{Proceedings of the IEEE international conference on computer
  vision}, pages 618--626, 2017.

\bibitem[Seyyed-Kalantari et~al.(2020)Seyyed-Kalantari, Liu, McDermott, Chen,
  and Ghassemi]{seyyedkalantari2020chexclusion}
Laleh Seyyed-Kalantari, Guanxiong Liu, Matthew McDermott, Irene~Y Chen, and
  Marzyeh Ghassemi.
\newblock Chexclusion: Fairness gaps in deep chest x-ray classifiers.
\newblock In \emph{BIOCOMPUTING 2021: Proceedings of the Pacific Symposium},
  pages 232--243. World Scientific, 2020.

\bibitem[Shamshad et~al.(2022)Shamshad, Khan, Zamir, Khan, Hayat, Khan, and
  Fu]{shamshad2022transformers}
Fahad Shamshad, Salman Khan, Syed~Waqas Zamir, Muhammad~Haris Khan, Munawar
  Hayat, Fahad~Shahbaz Khan, and Huazhu Fu.
\newblock Transformers in medical imaging: A survey.
\newblock \emph{arXiv preprint arXiv:2201.09873}, 2022.

\bibitem[Shen and Gao(2018)]{shen2018dynamic}
Yan Shen and Mingchen Gao.
\newblock Dynamic routing on deep neural network for thoracic disease
  classification and sensitive area localization.
\newblock In \emph{International Workshop on Machine Learning in Medical
  Imaging}, pages 389--397. Springer, 2018.

\bibitem[Shung et~al.(2012)Shung, Smith, and Tsui]{shung2012principles}
K~Kirk Shung, Michael Smith, and Benjamin~MW Tsui.
\newblock \emph{Principles of medical imaging}.
\newblock Academic Press, 2012.

\bibitem[Siewerdsen et~al.()Siewerdsen, Antonuk, El-Mohri, Yorkston, Huang,
  Boudry, and Cunningham]{siewerdsen1997empirical}
JH~Siewerdsen, LE~Antonuk, Y~El-Mohri, J~Yorkston, W~Huang, JM~Boudry, and
  IA~Cunningham.
\newblock Empirical and theoretical investigation of the noise performance of
  indirect detection, active matrix flat-panel imagers (amfpis) for diagnostic
  radiology.
\newblock \emph{Medical physics}.

\bibitem[Siewerdsen et~al.(1998)Siewerdsen, Antonuk, El-Mohri, Yorkston, Huang,
  and Cunningham]{siewerdsen1998signal}
JH~Siewerdsen, LE~Antonuk, Y~El-Mohri, J~Yorkston, W~Huang, and IA~Cunningham.
\newblock Signal, noise power spectrum, and detective quantum efficiency of
  indirect-detection flat-panel imagers for diagnostic radiology.
\newblock \emph{Medical physics}, 25\penalty0 (5):\penalty0 614--628, 1998.

\bibitem[Simpson et~al.(2019)Simpson, Antonelli, Bakas, Bilello, Farahani,
  Van~Ginneken, Kopp-Schneider, Landman, Litjens, Menze,
  et~al.]{simpson2019large}
Amber~L Simpson, Michela Antonelli, Spyridon Bakas, Michel Bilello, Keyvan
  Farahani, Bram Van~Ginneken, Annette Kopp-Schneider, Bennett~A Landman, Geert
  Litjens, Bjoern Menze, et~al.
\newblock A large annotated medical image dataset for the development and
  evaluation of segmentation algorithms.
\newblock \emph{arXiv preprint arXiv:1902.09063}, 2019.

\bibitem[Sourati et~al.(2019)Sourati, Gholipour, Dy, Tomas-Fernandez, Kurugol,
  and Warfield]{sourati2019intelligent}
Jamshid Sourati, Ali Gholipour, Jennifer~G Dy, Xavier Tomas-Fernandez, Sila
  Kurugol, and Simon~K Warfield.
\newblock Intelligent labeling based on fisher information for medical image
  segmentation using deep learning.
\newblock \emph{IEEE transactions on medical imaging}, 38\penalty0
  (11):\penalty0 2642--2653, 2019.

\bibitem[Sprawls(1993)]{sprawls1993physical}
Perry Sprawls.
\newblock Physical principles of medical imaging.
\newblock 1993.

\bibitem[Steiner et~al.(2021)Steiner, Kolesnikov, Zhai, Wightman, Uszkoreit,
  and Beyer]{steiner2021train}
Andreas Steiner, Alexander Kolesnikov, Xiaohua Zhai, Ross Wightman, Jakob
  Uszkoreit, and Lucas Beyer.
\newblock How to train your vit? data, augmentation, and regularization in
  vision transformers.
\newblock \emph{arXiv preprint arXiv:2106.10270}, 2021.

\bibitem[Suetens(2017)]{suetens2017fundamentals}
Paul Suetens.
\newblock \emph{Fundamentals of medical imaging}.
\newblock Cambridge university press, 2017.

\bibitem[Tang et~al.(2022)Tang, Yang, Li, Roth, Landman, Xu, Nath, and
  Hatamizadeh]{tang2022self}
Yucheng Tang, Dong Yang, Wenqi Li, Holger~R Roth, Bennett Landman, Daguang Xu,
  Vishwesh Nath, and Ali Hatamizadeh.
\newblock Self-supervised pre-training of swin transformers for 3d medical
  image analysis.
\newblock In \emph{Proceedings of the IEEE/CVF Conference on Computer Vision
  and Pattern Recognition}, pages 20730--20740, 2022.

\bibitem[Tang et~al.(2018)Tang, Wang, Harrison, Lu, Xiao, and
  Summers]{tang2018attention}
Yuxing Tang, Xiaosong Wang, Adam~P Harrison, Le~Lu, Jing Xiao, and Ronald~M
  Summers.
\newblock Attention-guided curriculum learning for weakly supervised
  classification and localization of thoracic diseases on chest radiographs.
\newblock In \emph{International Workshop on Machine Learning in Medical
  Imaging}, pages 249--258. Springer, 2018.

\bibitem[Taslimi et~al.(2022)Taslimi, Taslimi, Fathi, Salehi, and
  Rohban]{taslimi2022swinchex}
Sina Taslimi, Soroush Taslimi, Nima Fathi, Mohammadreza Salehi, and
  Mohammad~Hossein Rohban.
\newblock Swinchex: Multi-label classification on chest x-ray images with
  transformers.
\newblock \emph{arXiv preprint arXiv:2206.04246}, 2022.

\bibitem[Tay et~al.(2022)Tay, Dehghani, Abnar, Chung, Fedus, Rao, Narang, Tran,
  Yogatama, and Metzler]{tay2022scaling}
Yi~Tay, Mostafa Dehghani, Samira Abnar, Hyung~Won Chung, William Fedus, Jinfeng
  Rao, Sharan Narang, Vinh~Q Tran, Dani Yogatama, and Donald Metzler.
\newblock Scaling laws vs model architectures: How does inductive bias
  influence scaling?
\newblock \emph{arXiv preprint arXiv:2207.10551}, 2022.

\bibitem[Touvron et~al.(2021)Touvron, Cord, Douze, Massa, Sablayrolles, and
  J{\'e}gou]{touvron2021training}
Hugo Touvron, Matthieu Cord, Matthijs Douze, Francisco Massa, Alexandre
  Sablayrolles, and Herv{\'e} J{\'e}gou.
\newblock Training data-efficient image transformers \& distillation through
  attention.
\newblock In \emph{International Conference on Machine Learning}, pages
  10347--10357. PMLR, 2021.

\bibitem[Van~Ginneken et~al.(2001)Van~Ginneken, Romeny, and
  Viergever]{van2001computer}
Bram Van~Ginneken, BM~Ter~Haar Romeny, and Max~A Viergever.
\newblock Computer-aided diagnosis in chest radiography: a survey.
\newblock \emph{IEEE Transactions on medical imaging}, 20\penalty0
  (12):\penalty0 1228--1241, 2001.

\bibitem[Vaswani et~al.(2017)Vaswani, Shazeer, Parmar, Uszkoreit, Jones, Gomez,
  Kaiser, and Polosukhin]{vaswani2017attention}
Ashish Vaswani, Noam Shazeer, Niki Parmar, Jakob Uszkoreit, Llion Jones,
  Aidan~N Gomez, Lukasz Kaiser, and Illia Polosukhin.
\newblock Attention is all you need.
\newblock \emph{arXiv preprint arXiv:1706.03762}, 2017.

\bibitem[Wang et~al.(2019)Wang, Jia, Lu, and Xia]{wang2019thorax}
Hongyu Wang, Haozhe Jia, Le~Lu, and Yong Xia.
\newblock Thorax-net: an attention regularized deep neural network for
  classification of thoracic diseases on chest radiography.
\newblock \emph{IEEE journal of biomedical and health informatics}, 24\penalty0
  (2):\penalty0 475--485, 2019.

\bibitem[Wang et~al.(2020)Wang, Lin, and Wong]{Wang2020covid}
Linda Wang, Zhong~Qiu Lin, and Alexander Wong.
\newblock Covid-net: a tailored deep convolutional neural network design for
  detection of covid-19 cases from chest x-ray images.
\newblock \emph{Scientific Reports}, 10\penalty0 (1):\penalty0 19549, Nov 2020.
\newblock ISSN 2045-2322.
\newblock \doi{10.1038/s41598-020-76550-z}.
\newblock URL \url{https://doi.org/10.1038/s41598-020-76550-z}.

\bibitem[Wang et~al.(2017)Wang, Peng, Lu, Lu, Bagheri, and
  Summers]{wang2017chestx}
Xiaosong Wang, Yifan Peng, Le~Lu, Zhiyong Lu, Mohammadhadi Bagheri, and
  Ronald~M Summers.
\newblock Chestx-ray8: Hospital-scale chest x-ray database and benchmarks on
  weakly-supervised classification and localization of common thorax diseases.
\newblock In \emph{Proceedings of the IEEE conference on computer vision and
  pattern recognition}, pages 2097--2106, 2017.

\bibitem[Wightman(2019)]{rw2019timm}
Ross Wightman.
\newblock Pytorch image models.
\newblock \url{https://github.com/rwightman/pytorch-image-models}, 2019.

\bibitem[Xiang et~al.(2021)Xiang, Liu, Yuille, Zhang, Cai, and
  Zhou]{xiang2021painting}
Tiange Xiang, Yongyi Liu, Alan~L Yuille, Chaoyi Zhang, Weidong Cai, and Zongwei
  Zhou.
\newblock In-painting radiography images for unsupervised anomaly detection.
\newblock \emph{arXiv preprint arXiv:2111.13495}, 2021.

\bibitem[Xiao et~al.(2022)Xiao, Bai, Yuille, and Zhou]{xiao2022transforming}
Junfei Xiao, Yutong Bai, Alan Yuille, and Zongwei Zhou.
\newblock Transforming radiograph imaging with transformers: Comparing vision
  transformers with convolutional neural networks in multi-label thorax disease
  classification.
\newblock In \emph{Radiological Society of North America (RSNA)}, 2022.

\bibitem[Xiao et~al.(2023)Xiao, Bai, Yuille, and Zhou]{xiao2023delving}
Junfei Xiao, Yutong Bai, Alan Yuille, and Zongwei Zhou.
\newblock Delving into masked autoencoders for multi-label thorax disease
  classification.
\newblock In \emph{Proceedings of the IEEE/CVF Winter Conference on
  Applications of Computer Vision}, pages 3588--3600, 2023.

\bibitem[Xie et~al.(2021)Xie, Zhang, Shen, and Xia]{xie2021cotr}
Yutong Xie, Jianpeng Zhang, Chunhua Shen, and Yong Xia.
\newblock Cotr: Efficiently bridging cnn and transformer for 3d medical image
  segmentation.
\newblock In \emph{International conference on medical image computing and
  computer-assisted intervention}, pages 171--180. Springer, 2021.

\bibitem[Xie et~al.(2022)Xie, Zhang, Cao, Lin, Bao, Yao, Dai, and
  Hu]{xie2021simmim}
Zhenda Xie, Zheng Zhang, Yue Cao, Yutong Lin, Jianmin Bao, Zhuliang Yao,
  Qi~Dai, and Han Hu.
\newblock Simmim: A simple framework for masked image modeling.
\newblock In \emph{Proceedings of the IEEE/CVF Conference on Computer Vision
  and Pattern Recognition (CVPR)}, pages 9653--9663, June 2022.

\bibitem[Xu et~al.(2022)Xu, Zhu, and Wen]{xu2022deep}
Lanyu Xu, Simeng Zhu, and Ning Wen.
\newblock Deep reinforcement learning and its applications in medical imaging
  and radiation therapy: a survey.
\newblock \emph{Physics in Medicine \& Biology}, 67\penalty0 (22):\penalty0
  22TR02, 2022.

\bibitem[Yang and Yu(2021)]{yang2021artificial}
Ruixin Yang and Yingyan Yu.
\newblock Artificial convolutional neural network in object detection and
  semantic segmentation for medical imaging analysis.
\newblock \emph{Frontiers in oncology}, 11:\penalty0 638182, 2021.

\bibitem[Yao et~al.(2018)Yao, Prosky, Poblenz, Covington, and
  Lyman]{yao2018weakly}
Li~Yao, Jordan Prosky, Eric Poblenz, Ben Covington, and Kevin Lyman.
\newblock Weakly supervised medical diagnosis and localization from multiple
  resolutions.
\newblock \emph{arXiv preprint arXiv:1803.07703}, 2018.

\bibitem[Yao et~al.(2021)Yao, Liu, Zhou, Wang, Shen, Yuille, and
  Lu]{yao2021unsupervised}
Yuan Yao, Fengze Liu, Zongwei Zhou, Yan Wang, Wei Shen, Alan Yuille, and Yongyi
  Lu.
\newblock Unsupervised domain adaptation through shape modeling for medical
  image segmentation.
\newblock In \emph{Medical Imaging with Deep Learning}, 2021.

\bibitem[Yuan et~al.(2021)Yuan, Chen, Wang, Yu, Shi, Jiang, Tay, Feng, and
  Yan]{yuan2021tokens}
Li~Yuan, Yunpeng Chen, Tao Wang, Weihao Yu, Yujun Shi, Zi-Hang Jiang,
  Francis~EH Tay, Jiashi Feng, and Shuicheng Yan.
\newblock Tokens-to-token vit: Training vision transformers from scratch on
  imagenet.
\newblock In \emph{Proceedings of the IEEE/CVF International Conference on
  Computer Vision}, pages 558--567, 2021.

\bibitem[Zhang et~al.(2022)Zhang, Zhang, Zhang, Jin, Zhou, Cai, Zhao, Liu, and
  Liu]{zhang2022delving}
Chongzhi Zhang, Mingyuan Zhang, Shanghang Zhang, Daisheng Jin, Qiang Zhou,
  Zhongang Cai, Haiyu Zhao, Xianglong Liu, and Ziwei Liu.
\newblock Delving deep into the generalization of vision transformers under
  distribution shifts.
\newblock In \emph{Proceedings of the IEEE/CVF Conference on Computer Vision
  and Pattern Recognition}, pages 7277--7286, 2022.

\bibitem[Zhang et~al.(2018)Zhang, Ciss{\'{e}}, Dauphin, and
  Lopez{-}Paz]{ZhangCDL18}
Hongyi Zhang, Moustapha Ciss{\'{e}}, Yann~N. Dauphin, and David Lopez{-}Paz.
\newblock mixup: Beyond empirical risk minimization.
\newblock In \emph{ICLR}, 2018.

\bibitem[Zhou et~al.(2022{\natexlab{a}})Zhou, Yu, Xie, Xiao, Anandkumar, Feng,
  and Alvarez]{zhou2022understanding}
Daquan Zhou, Zhiding Yu, Enze Xie, Chaowei Xiao, Animashree Anandkumar, Jiashi
  Feng, and Jose~M Alvarez.
\newblock Understanding the robustness in vision transformers.
\newblock In \emph{International Conference on Machine Learning}, pages
  27378--27394. PMLR, 2022{\natexlab{a}}.

\bibitem[Zhou et~al.(2022{\natexlab{b}})Zhou, Wei, Wang, Shen, Xie, Yuille, and
  Kong]{zhou2021ibot}
Jinghao Zhou, Chen Wei, Huiyu Wang, Wei Shen, Cihang Xie, Alan Yuille, and Tao
  Kong.
\newblock ibot: Image bert pre-training with online tokenizer.
\newblock In \emph{ICLR}, 2022{\natexlab{b}}.

\bibitem[Zhou et~al.(2019{\natexlab{a}})Zhou, Rueckert, and
  Fichtinger]{zhou2019handbook}
S~Kevin Zhou, Daniel Rueckert, and Gabor Fichtinger.
\newblock \emph{Handbook of medical image computing and computer assisted
  intervention}.
\newblock Academic Press, 2019{\natexlab{a}}.

\bibitem[Zhou et~al.(2021{\natexlab{a}})Zhou, Le, Luu, Nguyen, and
  Ayache]{zhou2021deep}
S~Kevin Zhou, Hoang~Ngan Le, Khoa Luu, Hien~V Nguyen, and Nicholas Ayache.
\newblock Deep reinforcement learning in medical imaging: A literature review.
\newblock \emph{Medical image analysis}, 73:\penalty0 102193,
  2021{\natexlab{a}}.

\bibitem[Zhou et~al.(2012)Zhou, Zhan, Raykar, Hermosillo, Bogoni, and
  Peng]{zhou2012mining}
Xiang~Sean Zhou, Yiqiang Zhan, Vikas~C Raykar, Gerardo Hermosillo, Luca Bogoni,
  and Zhipang Peng.
\newblock Mining anatomical, physiological and pathological information from
  medical images.
\newblock \emph{ACM SIGKDD Explorations Newsletter}, 14\penalty0 (1):\penalty0
  25--34, 2012.

\bibitem[Zhou(2021)]{zhou2021towards}
Zongwei Zhou.
\newblock \emph{Towards Annotation-Efficient Deep Learning for Computer-Aided
  Diagnosis}.
\newblock PhD thesis, Arizona State University, 2021.

\bibitem[Zhou et~al.(2018)Zhou, Siddiquee, Tajbakhsh, and
  Liang]{zhou2018unet++}
Zongwei Zhou, Md~Mahfuzur~Rahman Siddiquee, Nima Tajbakhsh, and Jianming Liang.
\newblock Unet++: A nested u-net architecture for medical image segmentation.
\newblock In \emph{Deep Learning in Medical Image Analysis and Multimodal
  Learning for Clinical Decision Support}, pages 3--11. Springer, 2018.

\bibitem[Zhou et~al.(2019{\natexlab{b}})Zhou, Sodha, Siddiquee, Feng,
  Tajbakhsh, Gotway, and Liang]{zhou2019models}
Zongwei Zhou, Vatsal Sodha, Md~Mahfuzur~Rahman Siddiquee, Ruibin Feng, Nima
  Tajbakhsh, Michael~B Gotway, and Jianming Liang.
\newblock Models genesis: Generic autodidactic models for 3d medical image
  analysis.
\newblock In \emph{International conference on medical image computing and
  computer-assisted intervention}, pages 384--393. Springer,
  2019{\natexlab{b}}.

\bibitem[Zhou et~al.(2021{\natexlab{b}})Zhou, Sodha, Pang, Gotway, and
  Liang]{zhou2021models}
Zongwei Zhou, Vatsal Sodha, Jiaxuan Pang, Michael~B Gotway, and Jianming Liang.
\newblock Models genesis.
\newblock \emph{Medical image analysis}, 67:\penalty0 101840,
  2021{\natexlab{b}}.

\bibitem[Zhou et~al.(2022{\natexlab{c}})Zhou, Gotway, and
  Liang]{zhou2022interpreting}
Zongwei Zhou, Michael Gotway, and Jianming Liang.
\newblock Interpreting medical images.
\newblock In \emph{Intelligent Systems in Medicine and Health: The Role of AI}.
  Springer, 2022{\natexlab{c}}.

\bibitem[Zhu et~al.(2020)Zhu, Li, Hu, Ma, Zhou, and Zheng]{zhu2020rubik}
Jiuwen Zhu, Yuexiang Li, Yifan Hu, Kai Ma, S~Kevin Zhou, and Yefeng Zheng.
\newblock Rubik’s cube+: A self-supervised feature learning framework for 3d
  medical image analysis.
\newblock \emph{Medical Image Analysis}, 64:\penalty0 101746, 2020.

\bibitem[Zhu et~al.(2021)Zhu, Su, Lu, Li, Wang, and Dai]{zhu2020deformable}
Xizhou Zhu, Weijie Su, Lewei Lu, Bin Li, Xiaogang Wang, and Jifeng Dai.
\newblock Deformable detr: Deformable transformers for end-to-end object
  detection.
\newblock \emph{ICLR}, 2021.

\end{thebibliography}
